\documentclass{article}

     \PassOptionsToPackage{numbers, compress}{natbib}

    \usepackage[final]{neurips_2023}

\usepackage[utf8]{inputenc} 
\usepackage[T1]{fontenc}    
\usepackage[colorlinks=true,linkcolor=black,citecolor=blue,urlcolor=blue,]{hyperref}       
\usepackage{url}            
\usepackage{booktabs}       
\usepackage{amsfonts}       
\usepackage{nicefrac}       
\usepackage{microtype}      
\usepackage{xcolor}         
\usepackage{graphicx}
\usepackage{amsmath}
\usepackage{bm}
\usepackage{amssymb}
\usepackage{wrapfig}
\usepackage{subfigure}
\usepackage{algorithm}
\usepackage{algorithmic}
\usepackage{multirow}

\title{Efficient Meta Neural Heuristic for Multi-Objective Combinatorial Optimization}

\author{
Jinbiao Chen$^{1}$,
Jiahai Wang$^{1,2,3,}$\thanks{Jiahai Wang and Zizhen Zhang are the corresponding authors.}~~,
Zizhen Zhang$^{1,*}$,
Zhiguang Cao$^{4}$,
Te Ye$^{1}$,
Siyuan Chen$^{1}$\\
$^1$School of Computer Science and Engineering, Sun Yat-sen University, P.R. China\\
$^2$Key Laboratory of Machine Intelligence and Advanced Computing, Ministry of Education,\\Sun Yat-sen University, P.R. China\\
$^3$Guangdong Key Laboratory of Big Data Analysis and Processing, Guangzhou, P.R. China\\
$^4$School of Computing and Information Systems, Singapore Management University, Singapore\\
\texttt{chenjb69@mail2.sysu.edu.cn},~\texttt{\{wangjiah,zhangzzh7\}@mail.sysu.edu.cn}\\
\texttt{zgcao@smu.edu.sg},~\texttt{\{yete,chensy47\}@mail2.sysu.edu.cn}
}

\begin{document}

\maketitle

\begin{abstract}

Recently, neural heuristics based on deep reinforcement learning have exhibited promise in solving multi-objective combinatorial optimization problems (MOCOPs). However, they are still struggling to achieve high learning efficiency and solution quality. To tackle this issue, we propose an efficient meta neural heuristic (EMNH), in which a meta-model is first trained and then fine-tuned with a few steps to solve corresponding single-objective subproblems. Specifically, for the training process, a (partial) architecture-shared multi-task model is leveraged to achieve parallel learning for the meta-model, so as to speed up the training; meanwhile, a scaled symmetric sampling method with respect to the weight vectors is designed to stabilize the training. For the fine-tuning process, an efficient hierarchical method is proposed to systematically tackle all the subproblems. Experimental results on the multi-objective traveling salesman problem (MOTSP), multi-objective capacitated vehicle routing problem (MOCVRP), and multi-objective knapsack problem (MOKP) show that, EMNH is able to outperform the state-of-the-art neural heuristics in terms of solution quality and learning efficiency, and yield competitive solutions to the strong traditional heuristics while consuming much shorter time.

\end{abstract}

\section{Introduction}

Multi-objective combinatorial optimization problems (MOCOPs) \cite{ehr08} are widely studied and applied in many real-world sectors, such as telecommunication, logistics, manufacturing, and inventory. Typically, an MOCOP requires the simultaneous optimization of multiple conflicting objectives, where the amelioration of an objective may lead to the deterioration of others. Therefore, a set of trade-off solutions, known as \emph{Pareto-optimal} solutions, are usually sought for MOCOPs.

Generally, it is difficult to exactly find all the Pareto-optimal solutions of an MOCOP \cite{flo14}, especially given that the decomposed single-objective subproblem might already be NP-hard. Hence, heuristic methods \cite{her21} are usually preferred to solve MOCOPs in reality, as they can attain approximate Pareto-optimal solutions in (relatively) reasonable time. Nevertheless, the traditional heuristics are still lacking due to their reliance on handcrafted rules and massive iterative steps, and superfluous computation even for instances of the same (or similar) class.

Recently, inspired by the success of deep reinforcement learning (DRL) in learning neural heuristics for solving the single-objective combinatorial optimization problems (COPs) \cite{mir21,ben21,maz21,wan21}, a number of DRL-based neural heuristics \cite{lik21,wuh20,sha21,zha21,lin22,zha22} have also been investigated for MOCOPs. While bypassing the handcrafted rules, these neural heuristics adopt an end-to-end paradigm to construct solutions without iterative search. Benefiting from a large amount of data (i.e., problem instances), a well-trained deep model allows the neural heuristic to automatically extract informative and expressive features for decision making and generalize to unseen instances as well.

\begin{figure}[!t]
	\centering
	\includegraphics[width=\textwidth]{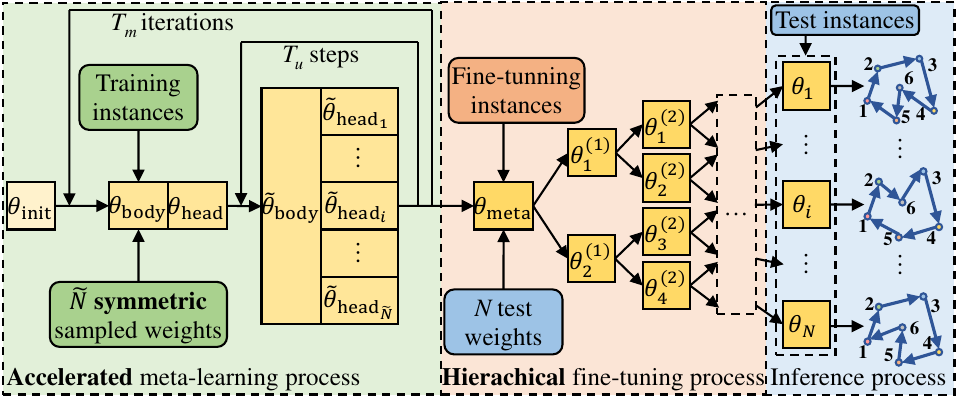}
	\caption{The overall framework of EMNH.}
	\label{fig1}
\end{figure}

Although demonstrating promise for solving MOCOPs, those neural heuristics still struggle to achieve high learning efficiency and solution quality. Particularly, in this line of works, the early attempts \cite{lik21,wuh20,sha21,zha21} always train multiple deep models, i.e., one for each preference (weight) combination, making them less practical. While the well-known preference-conditioned multi-objective combinatorial optimization (PMOCO) \cite{lin22} realized a unified deep model by introducing a huge hypernetwork, it leaves considerable gaps in terms of solution quality. The recent Meta-DRL (MDRL) \cite{zha22} has demonstrated the capability to enhance solution quality over existing state-of-the-art algorithms. However, it still faces challenges related to inefficient and unstable training procedures, as well as undesirable fine-tuning processes. This paper thereby proposes an efficient meta neural heuristic (EMNH) to further strengthen the learning efficiency and solution quality, the framework of which is illustrated in Figure \ref{fig1}. Following the meta-learning paradigm \cite{zha22,nic18}, EMNH first trains a meta-model and then quickly fine-tunes it according to the weight (preference) vector to tackle the corresponding single-objective subproblems.

Our contributions are summarized as follows. (1) We propose an efficient meta-learning scheme to accelerate the training. Inspired by the feature reuse of meta-learning~\cite{rag20}, we introduce a multi-task model composed of a parameter-shared \emph{body} (all other layers of neural network except the last one) and respective task-related \emph{heads} (the last layer of the network), rather than respective submodels, to train the meta-model, which is able to handle different tasks (subproblems) in parallel. (2) We design a scaled symmetric sampling method regarding the weight vectors to stabilize the training. In each meta-iteration, once a weight vector is sampled, its scaled symmetric weight vectors will also be engendered, which help avoid fluctuations of the parameter update for the meta-model, especially on problems with objectives of imbalanced domains. (3) We present a hierarchical fine-tuning method to efficiently cope with all the subproblems. Since it is redundant to derive a specific submodel for each subproblem in the early tuning process, our meta-model can be gradually fine-tuned to act as specialized submodels from low level to high level, which takes much fewer fine-tuning steps in total. Moreover, experimental results on three classic MOCOPs confirmed the effectiveness of our designs.

\section{Related works}

\textbf{Exact and heuristic methods for MOCOPs.} Over the past decades, MOCOPs have been attracting increasing attention from the computational intelligence community, and tremendous works have been proposed. Generally, exact methods \cite{flo14} are able to gain the accurate Pareto-optimal solutions, but their computational time may exponentially grow. Heuristic methods such as multi-objective evolutionary algorithms (MOEAs) are popular in practice, where the representative paradigms for the general multi-objective optimization include NSGA-II \cite{deb02} and MOEA/D \cite{zha07}. Among them, a number of MOEAs outfitted with local search \cite{jas02,shi20,shi22} are specifically designed for MOCOPs, which iteratively perform a search in the solution space to find approximate Pareto-optimal solutions.

\textbf{Neural heuristics for COPs.} As the early attempts, neural \emph{construction} heuristics based on deep models \cite{vin15,bel17,naz18} were proposed to learn to directly construct solutions for (single-objective) COPs. Among them, the classic \emph{Attention Model} (AM) \cite{koo19} was proposed inspired by \emph{Transformer} architecture \cite{vas17}, and known as a milestone for solving vehicle routing problems. A number of subsequent works \cite{kwo20,kim21,kwo21,xin21,pen20} were developed based on AM, including the popular policy optimization with multiple optima (POMO) \cite{kwo20}, which leveraged the solution symmetries and significantly enhanced the performance over AM. Besides, graph neural networks were also employed to learn graphic embeddings for solving COPs~\cite{kha17,man20,dua20,che22}. Different from the above neural construction heuristics, neural \emph{improvement} heuristics \cite{che19,luh20,hot20,wuy21,may21} were proposed to assist iteration-based methods to refine an initial solution.

\textbf{Neural heuristics for MOCOPs.} MOCOPs seek for a set of trade-off solutions, which could be obtained by solving a series of single-objective subproblems. According to the number of trained models, neural heuristics could be roughly classified into multi-model and single-model ones. The former adopts multiple networks to tackle respective subproblems, where the set of networks could also be collaboratively trained with a neighborhood-based parameter-transfer strategy \cite{lik21,wuh20}, or evolved with MOEA \cite{sha21,zha21}. Differently, the latter usually exploits a unified network to tackle all the subproblems, which is more flexible and practical. For example, taking the weight (preference) vector of a subproblem as the input, PMOCO introduces a hypernetwork to learn the decoder parameters \cite{lin22}. However, the hypernetwork may cause extra complexity to the original deep reinforcement learning model, rendering it less effective to learn more accurate mapping from the preference to the optimal solution of the corresponding subproblem. By contrast, MDRL \cite{zha22} leverages the meta-learning to train a deep reinforcement learning model that could be fine-tuned for various subproblems. However, its learning efficiency is far from optimal, especially given the slow and unstable training due to sequential learning of randomly sampled tasks, and inefficient fine-tuning due to redundant updates.

\section{Preliminary}

\subsection{MOCOP}

In general, an MOCOP could be defined as
\begin{equation}
	\underset{x \in \mathcal{X} }{\min}\; \bm{f}(x) = (f_1(x), f_2(x), \dots, f_M(x)),
\end{equation}
where $\bm{f}(x)$ is an $M$-objective vector, $\mathcal{X}$ is a discrete decision space, and the objectives might be conflicted. The Pareto-optimal solutions concerned by decision makers are defined as follows.

\textbf{Definition 1 (Pareto dominance).} Let $u, v \in \mathcal{X} $, $u$ is said to dominate $v$ ($ u \prec v$) if and only if $f_i(u) \leq f_i(v), \forall i \in \{1, \dots, M\}$ and $\exists j \in \{1, \dots, M\}, f_j(u) < f_j(v) $.

\textbf{Definition 2 (Pareto optimality).} A solution $x^* \in \mathcal{X}$ is Pareto-optimal if $x^*$ is not dominated by any other solution in $\mathcal{X}$, i.e., $\nexists x' \in \mathcal{X}$ such that $x' \prec x^*$. The set of all Pareto optimal solutions $\mathcal{P} = \{x^* \in \mathcal{X}~|~ \nexists x' \in \mathcal{X}: x' \prec x^* \}$ is called the Pareto set. The \emph{image} of Pareto set in the objective space, i.e., $\mathcal{PF} = \{\bm{f}(x) ~|~ x \in \mathcal{P} \}$ is called the Pareto front.

\subsection{Decomposition}

The decomposition strategy is always applied for solving MOCOPs~\cite{zha07}, since it is straightforward yet effective to guide the search towards prescribed directions. Specifically, MOCOPs can be decomposed into $N$ scalarized single-objective subproblem $g(x|\bm{\lambda})$, where the weight vector $\bm{\lambda}\in \mathcal{R}^M$ satisfies $\lambda_m\geq0$ and $\sum_{m=1}^{M}\lambda_m=1$. Then, $\mathcal{PF}$ is approximated by solving the subproblems systematically.

Regarding the decomposition strategy, the weighted sum (WS) and Tchebycheff \cite{max17} are commonly used. As a simple representative, WS considers the linear combination of $M$ objectives as follows,
\begin{equation}
	\underset{x \in \mathcal{X}}{\min}\; g_{\rm{ws}}(x|\bm{\lambda}) = \sum_{m=1}^{M}\lambda_m f_m(x).
\end{equation}

Given the weight vector $\bm{\lambda}$, the corresponding single-objective subproblem could be cast as a sequential decision problem and solved through DRL. In particular, a solution is represented as a sequence $\bm{\pi}=\{\pi_1,\dots,\pi_{T}\}$ with length $T$, and a stochastic policy for yielding solution $\bm{\pi}$ from instance $s$ is calculated as $P(\bm{\pi}|s)=\prod_{t=1}^{T} P_{\bm{\theta}}(\pi_t|\bm{\pi}_{1:t-1},s)$, where $P_{\bm{\theta}}(\pi_t|\bm{\pi}_{1:t-1},s)$ is the node selection probability parameterized by $\bm{\theta}$ (the subscript of $\bm{\theta}_i$, $\bm{\pi}_i$, $P_i$ for subproblem $i$ is omitted for readability).

\subsection{Meta-learning}

Meta-learning \cite{nic18,fin17} aims to train a model that can learn to tackle various tasks (i.e., subproblems) via fine-tuning. Naturally, a meta-model can be trained to quickly adapt to $N$ corresponding subproblems given $N$ weight vectors. Generally, meta-learning comprises three processes for MOCOPs. In the meta-learning process, a model $\bm{\theta}$ is trained by repeatedly sampling tasks from the whole task space. In the fine-tuning process, $N$ submodels $\bm{\theta}_1, \dots, \bm{\theta}_N$ based on the given weight vectors are fine-tuned from $\bm{\theta}$ using fine-tuning instances (also called query set). In the inference process, $N$ subproblems of an instance are solved using the submodels to approximate the Pareto set.

\section{Methodology}

Our \emph{efficient meta neural heuristic} (EMNH) includes an accelerated and stabilized meta-learning process, a hierarchical fine-tuning process, and an inference process, as depicted in Figure \ref{fig1}. EMNH is generic, where we adopt the first-order gradient-based \emph{Reptile} \cite{nic18} as the backbone of meta-learning, and employ the popular neural solver POMO \cite{kwo20} as the base model. In our meta-learning process, the model $\bm{\theta}$ is trained with $T_m$ meta-iterations, where a multi-task model is exploited to speed up the training and a scaled symmetric sampling method is designed to stabilize the training. In our fine-tuning process, the meta-model $\bm{\theta}$ is hierarchically and quickly fine-tuned with a few gradient updates to solve the corresponding subproblems. The details for each design are presented below.

\subsection{Accelerated training}

In the accelerated training process, as shown in Algorithm \ref{alg1}, the meta-model $\bm{\theta}$ is trained with $T_m$ meta-iterations. In each iteration, the model needs to be optimized by learning $\tilde{N}$ sampled tasks (subproblems) with $T_u$-step gradient updates. Inspired by the feature reuse of meta-learning \cite{rag20}, it is unnecessary and expensive to serially update the meta-model through the $\tilde{N}$ respective submodels of the same architecture. On the other hand, it is reasonable to assume that only the \emph{head} $\bm{\theta}_{\rm{head}}$ is specified for a task, while the \emph{body} $\bm{\theta}_{\rm{body}}$ can be reused for all tasks. To realize such a lightweight meta-learning scheme, we further introduce a multi-task model $\bm{\tilde{\theta}}$ which is composed of a shared body and $\tilde{N}$ respective heads to learn the $\tilde{N}$ tasks in parallel.

\begin{algorithm}[!t]
	\renewcommand{\algorithmicrequire}{\textbf{Input:}}
	\renewcommand{\algorithmicensure}{\textbf{Output:}}
	\caption{Accelerated training process}
	\label{alg1}
	\begin{algorithmic}[1]
		\REQUIRE initialized meta-model \bm{$\theta$}, number of symmetric sampled weight vectors $\tilde{N}$, initialized multi-task model \bm{$\tilde{\theta}$}, initial meta-learning rate $\epsilon_0$, number of meta-iterations $T_m$, number of update steps of the multi-task model $T_u$, batch size $B$, problem size $n$
		\STATE $\epsilon \leftarrow \epsilon_0$
		\FOR {$t_m = 1$ to $T_m$}
		\STATE $\bm{\lambda}_i$ is obtained by the scaled symmetric sampling method, $\quad \forall i \in \{1, \dots, \tilde{N}\}$
		\STATE $\bm{\tilde{\theta}}_{\rm{body}} \leftarrow \bm{\theta}_{\rm{body}}$
		\STATE $\bm{\tilde{\theta}}_{{\rm{head}}_i} \leftarrow \bm{\theta}_{\rm{head}}$, $\quad \forall i$
		\FOR {$t_u = 1$ to $T_u$}
		\STATE $s_j \sim \textbf{SampleInstance}(\mathcal{S})$, $\quad \forall j \in \{1, \dots, B\}$
		\STATE $\{\bm{\pi}^k|s_j,\bm{\lambda}_i\} \sim \textbf{SampleRollout}(P_{\bm{\tilde{\theta}}i}(\cdot|s_j))$, $\quad \forall k \in \{1, \dots, n\}$, $\forall i$, $\forall j$
		\STATE $b_{ij} \leftarrow \frac{1}{n}\sum_{k=1}^n g(\bm{\pi}^k|s_j,\bm{\lambda}_i)$
		\STATE $\nabla\mathcal{L}(\bm{\tilde{\theta}}) \leftarrow \frac{1}{\tilde{N}Bn}\sum\limits_{i=1}^{\tilde{N}}\sum\limits_{j=1}^B\sum\limits_{k=1}^n[(g_{ij}^k-b_{ij})\nabla\text{log}P_{\bm{\tilde{\theta}}i}(\bm{\pi}^k|s_j)]$
		\STATE $\bm{\tilde{\theta}} \leftarrow \textbf{Adam}(\bm{\tilde{\theta}}, \nabla\mathcal{L}(\bm{\tilde{\theta}}))$
		\ENDFOR
		\STATE $\bm{\theta}_{\rm{body}} \leftarrow \bm{\tilde{\theta}}_{\rm{body}}$
		\STATE $\bm{\theta}_{\rm{head}} \leftarrow \bm{\theta}_{\rm{head}}+\epsilon(\frac{1}{\tilde{N}}\sum_{i=1}^{\tilde{N}} \bm{\tilde{\theta}}_{{\rm{head}}_i}-\bm{\theta}_{\rm{head}})$
		\STATE $\epsilon \leftarrow \epsilon-\epsilon_0/T_m$
		\ENDFOR 
		\ENSURE The trained meta-model $\bm{\theta}$
	\end{algorithmic}  
\end{algorithm}

\begin{figure}[!t]
	\centering
	\subfigure[]{
			\centering
			\includegraphics[width=0.46\textwidth]{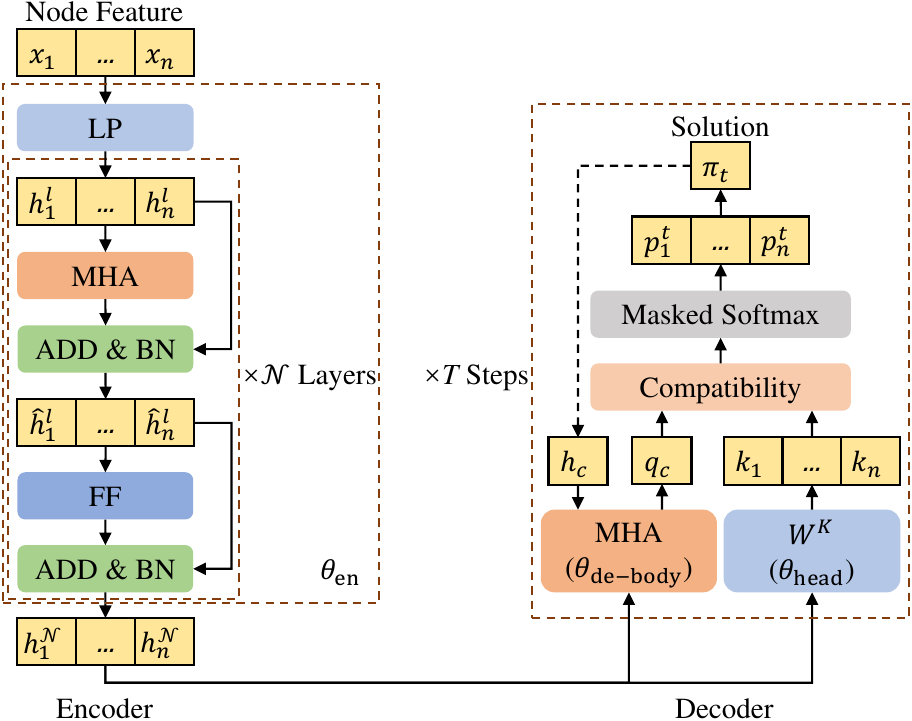}
			\label{figA1a}
		}
	\subfigure[]{
			\centering
			\includegraphics[width=0.50\textwidth]{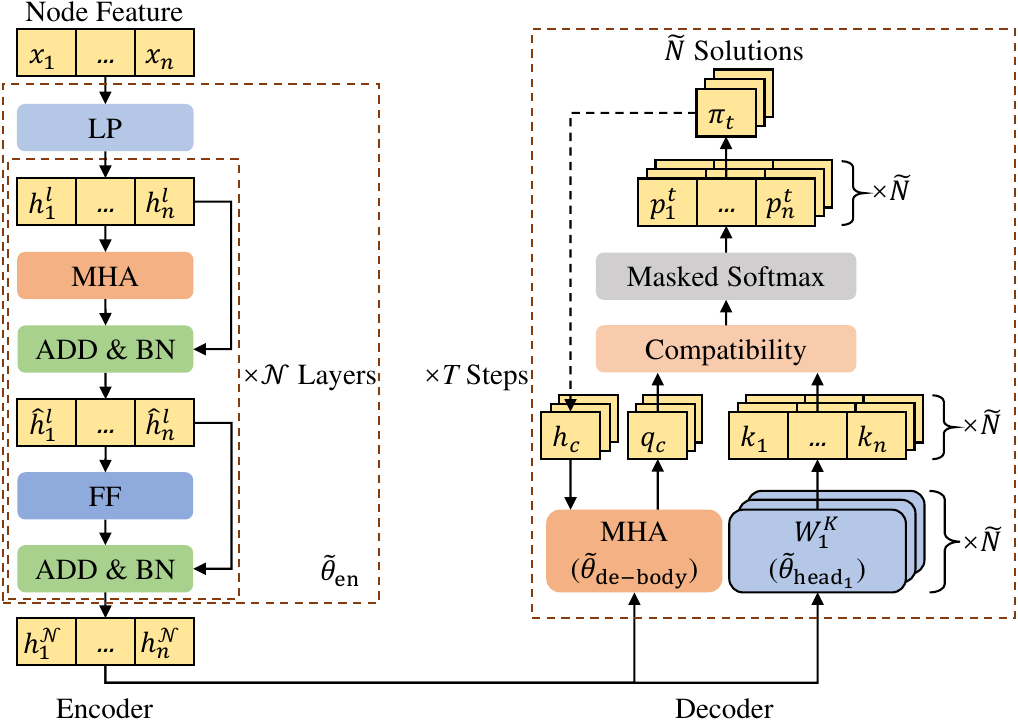}
			\label{figA1b}
		}
	\caption{Model architectures. (a) Base model (POMO). (b) Multi-task model.}
	\label{figA1}
\end{figure}

Specifically, since our EMNH adopts the encoder-decoder structured POMO as the base model, as depicted in Figure \ref{figA1a}, 1) $\bm{\theta}_{\rm{head}}$ could be defined as the decoder head, i.e., $W^K\in \mathcal{R}^{d\times d}$ in the last single-head attention layer, where $d$ is empirically set to 128 \cite{kwo20}; 2) $\bm{\theta}_{\rm{body}}$ could be composed of the whole encoder $\bm{\theta}_{\rm{en}}$ and the decoder body $\bm{\theta}_{\rm{de-body}}$. For a problem instance with $n$ nodes, the node embeddings $\bm{h}_1, \dots, \bm{h}_n \in \mathcal{R}^d$ are computed by $\bm{\theta}_{\rm{en}}$ at the encoding step. At each decoding step, the \emph{query} $\bm{q}_c \in \mathcal{R}^d$ is first computed by $\bm{\theta}_{\rm{de-body}}$ using the node embeddings and problem-specific \emph{context} embedding $\bm{h}_c$. Then, the last single-head attention layer computes the probability of node selection $P_{\bm{\theta}}(\bm{\pi}|s)$ using $\bm{q}_c$ and the \emph{key} $\bm{k}_1, \dots, \bm{k}_n \in \mathcal{R}^d$, where $\bm{k}_{i'}$ for node $i'$ is computed by $\bm{\theta}_{\rm{head}}$, i.e., $\bm{k}_{i'} = W^K \bm{h}_{i'}$. More details are presented in Appendix A.

As demonstrated in Figure \ref{figA1b}, for natural accommodation, the multi-task model $\bm{\tilde{\theta}}$ consists of $\bm{\tilde{\theta}}_{\rm{body}}$ and $\bm{\tilde{\theta}}_{{\rm{head}}_1}, \dots, \bm{\tilde{\theta}}_{{\rm{head}}_{\tilde{N}}}$, where $\bm{\tilde{\theta}}_{\rm{body}}$ and $\bm{\tilde{\theta}}_{{\rm{head}}_i}$ have the same architecture as $\bm{\theta}_{\rm{body}}$ and $\bm{\theta}_{\rm{head}}$, respectively. Note that $\bm{\tilde{\theta}}_{{\rm{head}}_i}$ is individually updated for subproblem $i$, while $\bm{\tilde{\theta}}_{\rm{body}}$ is shared for $\tilde{N}$ tasks. Concretely, since the subproblems are captured by different weight vectors but the same (or similar) node features, the shared node embeddings are computed by $\bm{\tilde{\theta}}_{\rm{en}}$. At each decoding step, for subproblem $i$, the \emph{query} $\bm{q}_{c,i}$ is first computed by $\bm{\tilde{\theta}}_{\rm{de-body}}$ using the node embeddings and \emph{context} embedding $\bm{h}_{c,i}$. Then, the \emph{key} is computed as $\bm{k}_{i',i} = W^K_{i} \bm{h}_{i'}$ for node $i'$. Finally, the single-head attention layer computes the probability of node selection $P_{\bm{\tilde{\theta}}i}(\bm{\pi}|s)$ for subproblem $i$ using $\bm{q}_{c,i}$ and $\bm{k}_{1,i}, \dots, \bm{k}_{n,i}$. In practice, $W^K_1, \dots, W^K_{\tilde{N}}$ are concatenated to tackle the $\tilde{N}$ tasks in parallel.

We extend REINFORCE \cite{wil92} to train the multi-task model $\bm{\tilde{\theta}}$, including $\bm{\tilde{\theta}}_{\rm{body}}$ for all tasks and $\bm{\tilde{\theta}}_{{\rm{head}}_i}$ for task $i$. Specifically, $\bm{\tilde{\theta}}$ is optimized by gradient descent with the multi-task loss $\nabla\mathcal{L}=\frac{1}{\tilde{N}}\sum_{i=1}^{\tilde{N}} \nabla\mathcal{L}_i$, where $\mathcal{L}_i(\bm{\tilde{\theta}}|s)$ is the $i$-th task loss and its gradient is estimated as follows,
\begin{equation}
	\nabla\mathcal{L}_i(\bm{\tilde{\theta}}|s)=\textbf{E}_{P_{\bm{\tilde{\theta}}i}(\bm{\pi}|s)}[(g(\bm{\pi}|s,\bm{\lambda}_i)-b_i(s))\nabla\text{log}P_{\bm{\tilde{\theta}}i}(\bm{\pi}|s)],
\end{equation}
where $b_i(s)=\frac{1}{n}\sum_{k=1}^n g(\bm{\pi}^k|s,\bm{\lambda}_i)$ is a baseline to reduce the gradient variance; $\bm{\pi}^1, \dots, \bm{\pi}^n$ are solutions produced by POMO with $n$ different starting nodes. In practice, $\nabla\mathcal{L}$ is approximated using Monte Carlo sampling, and computed by a batch with $B$ instances as follows,
\begin{equation}
	\nabla\mathcal{L}(\bm{\tilde{\theta}}) \approx \frac{1}{\tilde{N}Bn}\sum_{i=1}^{\tilde{N}}\sum_{j=1}^B\sum_{k=1}^n[(g_{ij}^k-b_{ij})\nabla\text{log}P_{\bm{\tilde{\theta}}i}(\bm{\pi}^k|s_j)],
\end{equation}
where $g_{ij}^k=g(\bm{\pi}^k|s_j,\bm{\lambda}_i)$ and $b_{ij}=b_i(s_j)$.  $\bm{\tilde{\theta}}$ is then optimized with $T_u$ steps using Adam \cite{kin15}.

\subsection{Stabilized training}

\begin{figure}[!t]
	\centering
	\includegraphics[width=0.7\textwidth]{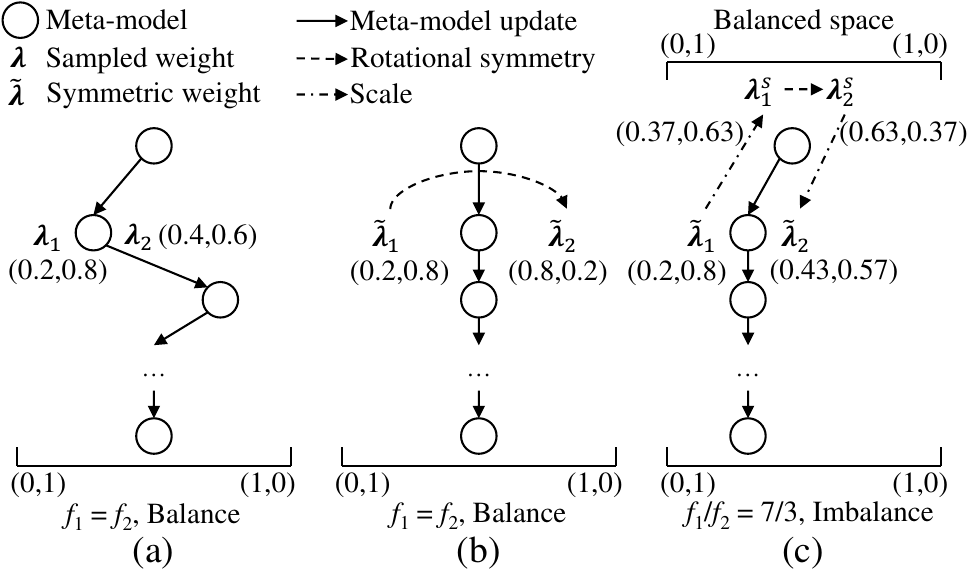} 
 \caption{ (a) Random sampling. (b) Symmetric sampling. (c) Scaled symmetric sampling.}
	\label{fig2}
\end{figure}

In each meta-iteration, the deviation of $\tilde{N}$ (usually a small number) randomly sampled weight vectors may cause fluctuations to the parameter update of the meta-model, thereby leading to unstable training performance. For example, as shown in Figure \ref{fig2}(a), the biased samples of weight vectors, i.e., $\bm{\lambda}_1$ and $\bm{\lambda}_2$, make the meta-model tend to favor the second objective more in this iteration.

To address this issue, we first design a symmetric sampling method for the case of balanced objective domain, as illustrated in Figure \ref{fig2}(b). Once a weight vector $\bm{\tilde{\lambda}}_1$ is sampled, its $M$-1 rotational symmetric weight vectors $\bm{\tilde{\lambda}}_2, \dots, \bm{\tilde{\lambda}}_M$ are calculated and used to reduce the bias as follows,

\begin{equation}
	\tilde{\lambda}_{i,m}=\left\{
	\begin{array}{lcl}
		\tilde{\lambda}_{i-1,M}, & & m=1\\
		\tilde{\lambda}_{i-1,m-1}, & & \rm{otherwise}
	\end{array} \right..
	\label{euq5}
\end{equation}

For the case of imbalanced objective domains, we further design a scaled symmetric sampling method. In particular, the $m$-th weight $\lambda_m$ is scaled as $\lambda_m/f^*_m$, where $f^*_m$ is the ideal objective value but difficult to be calculated beforehand. Thus, $f^*_m$ is replaced with $f'_m$, which is dynamically estimated using the meta-model on a validation dataset with $\bm{\lambda}=(1/M,\dots,1/M)$ during training.

Concretely, for a sampled $\bm{\tilde{\lambda}}_1$, it is first multiplied by $\bm{f}'$ to scale objectives into the balanced domains. Then, its $M$-1 rotational symmetric weight vectors are obtained by Eq. (\ref{euq5}). Finally, they are divided by $\bm{f}'$ to scale back to the original domain to attain the genuine rotational symmetric weight vectors. An example for $\bm{\tilde{\lambda}}_1=(0.2,0.8)$ and its scaled symmetric $\bm{\tilde{\lambda}}_2=(0.43,0.57)$ are demonstrated in Figure \ref{fig2}(c). In summary, the scaled symmetric $\bm{\tilde{\lambda}}_2, \dots, \bm{\tilde{\lambda}}_M$ for a given $\bm{\tilde{\lambda}}_1$ are calculated as follows,
\begin{equation}
	\lambda^s_{i,m}=\left\{
	\begin{array}{lcl}
		\tilde{\lambda}_{i-1,M}\times f'_M/f'_m, & & m=1\\
		\tilde{\lambda}_{i-1,m-1}\times f'_{m-1}/f'_m, & & \rm{otherwise}
	\end{array} \right..
\end{equation}
Then, the vectors are normalized as $\tilde{\lambda}_{i,m} = \frac{\lambda^s_{i,m}}{\sum_{m=1}^{M} \lambda^s_{i,m}}$. The pseudo code of the scaled symmetric sampling method for the weight vectors is presented in Appendix B.

\subsection{Hierarchical fine-tuning}

Once the meta-model is trained, it can be fine-tuned with a number of gradient updates to derive customized submodels for given weight vectors. However, it still might be expensive for MOCOPs since numerous submodels need to be individually fine-tuned from the meta-model. From a systematic perspective, the early coarse-tuning processes in which the parameters of neighboring submodels might be close, could be merged. In this sense, we propose a hierarchical method to efficiently fine-tune the numerous subproblems from the lowest to the highest level, as illustrated in Figure \ref{fig1}.

We suggest an $L$-level and $a$-section hierarchy to gradually produce more specific submodels. Particularly, all $N^{(l)}$ weight vectors in level $l$ are defined as the centers of their corresponding subspaces that uniformly divide the whole weight vector space based on the \citet{das98} method. $N^{(l)}$ submodels are then fine-tuned to derive $N^{(l+1)}=aN^{(l)}$ submodels with $K^{(l+1)}$ fine-tuning steps in level $l$+1. Finally, $N^{(L-1)}$ submodels are fine-tuned to derive $N^{(L)}=N$ submodels, which are used to solve the $N$ given subproblems.  More details of this fine-tuning are presented in Appendix C.

For the $L$-level and $a$-section hierarchy with $a^L=N$ and $K^{(1)}=\dots=K^{(L)}=K$, the total fine-tuning step is $T_f^h=\sum_{l=1}^{L}Ka^l=Ka(a^L-1)/(a-1)$. For the vanilla process, each submodel needs $KL$ fine-tuning steps to achieve comparable performance, thus $T_f^v=KLN=KLa^L$ steps in total. Hence, the hierarchical fine-tuning only needs $T_f^h/T_f^v\approx1/L$ of the vanilla fine-tuning steps.

\section{Experimental results}

\subsection{Experimental settings}

We conduct computational experiments to evaluate the proposed method on the multi-objective traveling salesman problem (MOTSP), multi-objective capacitated vehicle routing problem (MOCVRP), and multi-objective knapsack problem (MOKP). Following the convention in \cite{kwo20,lin22}, we consider the instances of different sizes $n$=20/50/100 for MOTSP/MOCVRP and $n$=50/100/200 for MOKP. All experiments are run on a PC with an Intel Xeon 4216 CPU and an RTX 3090 GPU.

\textbf{Problems.} Four classes of MOTSP \cite{lus10} are considered, including bi-objective TSP type 1\&2 (Bi-TSP-1, Bi-TSP-2) and tri-objective TSP type 1\&2 (Tri-TSP-1, Tri-TSP-2). For the $M$-objective TSP type 1, node $i$ has $M$ 2D coordinates $\{\bm{x}^1_i, \dots, \bm{x}^M_i\}$, where the $m$-th cost between node $i$ and $j$ is defined as the Euclidean distance $c^m_{ij} = \Vert \bm{x}^m_i - \bm{x}^m_j \Vert_2$. For the $M$-objective TSP type 2, a node contains $M$-1 2D coordinates together with the altitude, so the $M$-th cost is defined as the altitude variance. For the bi-objective CVRP (Bi-CVRP) \cite{joz08}, two conflicting objectives with imbalanced domains, i.e., the total traveling distance and the traveling distance of the longest route (also called makespan), are considered. For the bi-objective KP (Bi-KP) \cite{baz09}, item $i$ has a single weight and two distinct values. More details of these MOCOPs are presented in Appendix D.

\textbf{Hyper-parameters.} The meta-learning rate $\epsilon$ is linearly annealed to $0$ from $\epsilon_0 = 1$ initially. A constant learning rate of the Adam optimizer is set to $10^{-4}$. We set $B=64$, $T_m=3000$, $T_u=100$, and $\tilde{N}=M$. The $N$ weight vectors in $\mathcal{PF}$ construction are produced according to \cite{das98}, where $N=C^{M-1}_{H+M-1}$. $H$ is set to $100$ ($N=101$) and $13$ ($N=105$) for $M=2$ and $M=3$, respectively. WS scalarization is adopted. In the hierarchy of fine-tuning, the whole weight space is uniformly divided by the \citet{das98} method with $H^{(l)}=2^l$, i.e., level $l$ has $2^l$ or $4^l$ subspaces and $L=7$ or $L=4$ for $M=2$ or $M=3$. The number of fine-tuning steps at each level is $K^{(l)}=K=20$ or $K^{(l)}=K=25$ for $M=2$ or $M=3$.

\textbf{Baselines.} Two kinds of strong baselines are introduced, and all of them adopt WS (weighted sum) scalarization for fair comparisons. (1) The state-of-the-art neural heuristics, including \textbf{MDRL} \cite{zha22}, \textbf{PMOCO} \cite{lin22}, and DRL-based multiobjective optimization algorithm (\textbf{DRL-MOA}) \cite{lik21}. Same as EMNH, all these neural heuristics adopt POMO as the base model for single-objective subproblems. (2) The state-of-the-art traditional heuristics, including \textbf{PPLS/D-C} \cite{shi22}, \textbf{WS-LKH}, and \textbf{WS-DP}. PPLS/D-C is a local-search-based MOEA proposed for MOCOPs, where a 2-opt heuristic is used for MOTSP and MOCVRP, and a greedy transformation heuristic \cite{ish15} is used for MOKP, running with 200 iterations for all the three problems. WS-LKH and WS-DP are based on the state-of-the-art heuristics for decomposed single-objective subproblems, i.e., LKH \cite{hel00,tin18} for MOTSP and dynamic programming (DP) for MOKP, respectively. Our code is publicly available\footnote{\url{https://github.com/bill-cjb/EMNH}}.

\textbf{Metrics.} Our EMNH is evaluated in terms of solution quality and learning efficiency. Solution quality is mainly measured by hypervolume (HV) \cite{aud21}, where a higher HV means a better solution set (the definition is presented in Appendix E). Learning efficiency mainly refers to the training and fine-tuning efficiency, which are measured by training time for the same amount of training instances and solution quality with the same total fine-tuning steps (see Appendix F), respectively.

\subsection{Solution quality}

The results for MOTSP, MOCVRP, and MOKP are recorded in Tables \ref{tab1} and \ref{tab2}, including the average HV, gap, and total running time for solving 200 random test instances. To further show the significant differences between the results, a Wilcoxon rank-sum test at 1\% significance level is adopted. The best result and the one without statistical significance to it are highlighted as \textbf{bold}. The second-best result and the one without statistical significance to it are highlighted as \underline{underline}. The method with ``-Aug" represents the inference results using instance augmentation \cite{lin22} (see Appendix G). 

\begin{table}[!t]
	\caption{Results on 200 random instances for MOCOPs with balanced objective domains.}
	\centering
    \addtolength{\tabcolsep}{-1pt}
	\begin{tabular}{l|ccc|ccc|ccc}
		\toprule
		& \multicolumn{3}{c|}{Bi-TSP-1 ($n$=20)} & \multicolumn{3}{c|}{Bi-TSP-1 ($n$=50)} & \multicolumn{3}{c}{Bi-TSP-1 ($n$=100)} \\
		Method & HV$\uparrow$    & Gap$\downarrow$   & Time  & HV$\uparrow$    & Gap$\downarrow$   & Time  & HV$\uparrow$    & Gap$\downarrow$   & Time \\
		\midrule
		WS-LKH & \underline{0.6270}  & \underline{0.02\%} & 10m   & \textbf{0.6415 } & \textbf{-0.11\%} & 1.8h  & \textbf{0.7090 } & \textbf{-0.95\%} & 6.0h \\
		PPLS/D-C & 0.6256  & 0.24\% & 26m   & 0.6282  & 1.97\% & 2.8h  & 0.6844  & 2.55\% & 11h \\
		\midrule
		DRL-MOA & 0.6257  & 0.22\% & 6s    & 0.6360  & 0.75\% & 9s    & 0.6970  & 0.75\% & 21s \\
		PMOCO & 0.6259  & 0.19\% & 6s    & 0.6351  & 0.89\% & 10s   & 0.6957  & 0.94\% & 19s \\
		MDRL  & \textbf{0.6271 } & \textbf{0.00\%} & 5s    & 0.6364  & 0.69\% & 9s    & 0.6969  & 0.77\% & 17s \\
		EMNH & \textbf{0.6271 } & \textbf{0.00\%} & 5s    & 0.6364  & 0.69\% & 9s    & 0.6969  & 0.77\% & 16s \\
		PMOCO-Aug & \underline{0.6270}  & \underline{0.02\%} & 45s   & 0.6395  & 0.20\% & 2.3m  & 0.7016  & 0.10\% & 15m \\
		MDRL-Aug & \textbf{0.6271 } & \textbf{0.00\%} & 33s   & \underline{0.6408}  & \underline{0.00\%} & 1.7m  & \underline{0.7022}  & \underline{0.01\%} & 14m \\
		EMNH-Aug & \textbf{0.6271 } & \textbf{0.00\%} & 33s   & \underline{0.6408}  & \underline{0.00\%} & 1.7m  & \underline{0.7023}  & \underline{0.00\%} & 14m \\
		\midrule
		& \multicolumn{3}{c|}{Bi-KP ($n$=50)} & \multicolumn{3}{c|}{Bi-KP ($n$=100)} & \multicolumn{3}{c}{Bi-KP ($n$=200)} \\
		Method & HV$\uparrow$    & Gap$\downarrow$   & Time  & HV$\uparrow$    & Gap$\downarrow$   & Time  & HV$\uparrow$    & Gap$\downarrow$   & Time \\
		\midrule
		WS-DP & \textbf{0.3561 } & \textbf{0.00\%} & 22m   & \underline{0.4532}  & \underline{0.07\%} & 2.0h  & \underline{0.3601}  & \underline{0.06\%} & 5.8h \\
		PPLS/D-C & 0.3528  & 0.93\% & 18m   & 0.4480  & 1.21\% & 47m   & 0.3541  & 1.72\% & 1.5h \\
		\midrule
		DRL-MOA & \underline{0.3559}  & \underline{0.06\%} & 9s    & 0.4531  & 0.09\% & 18s   & \underline{0.3601}  & \underline{0.06\%} & 1.0m \\
		PMOCO & 0.3552  & 0.25\% & 6s    & 0.4523  & 0.26\% & 22s   & 0.3595  & 0.22\% & 1.3m \\
		MDRL  & 0.3530  & 0.87\% & 6s    & \underline{0.4532}  & \underline{0.07\%} & 21s   & \underline{0.3601}  & \underline{0.06\%} & 1.2m \\
		EMNH & \textbf{0.3561 } & \textbf{0.00\%} & 6s    & \textbf{0.4535 } & \textbf{0.00\%} & 21s   & \textbf{0.3603 } & \textbf{0.00\%} & 1.2m \\
		\midrule
		& \multicolumn{3}{c|}{Tri-TSP-1 ($n$=20)} & \multicolumn{3}{c|}{Tri-TSP-1 ($n$=50)} & \multicolumn{3}{c}{Tri-TSP-1 ($n$=100)} \\
		Method & HV$\uparrow$    & Gap$\downarrow$   & Time  & HV$\uparrow$    & Gap$\downarrow$   & Time  & HV$\uparrow$    & Gap$\downarrow$   & Time \\
		\midrule
		WS-LKH & \textbf{0.4712 } & \textbf{0.00\%} & 12m   & \textbf{0.4440 } & \textbf{-0.50\%} & 1.9h  & \textbf{0.5076 } & \textbf{-2.07\%} & 6.6h \\
		PPLS/D-C & 0.4698  & 0.30\% & 1.4h  & 0.4174  & 5.52\% & 3.9h  & 0.4376  & 12.00\% & 14h \\
		\midrule
		DRL-MOA & \underline{0.4699}  & \underline{0.28\%} & 6s    & 0.4303  & 2.60\% & 9s    & 0.4806  & 3.36\% & 19s \\
		PMOCO & 0.4693  & 0.40\% & 5s    & 0.4315  & 2.33\% & 8s    & 0.4858  & 2.31\% & 18s \\
		MDRL  & \underline{0.4699}  & \underline{0.28\%} & 5s    & 0.4317  & 2.29\% & 9s    & 0.4852  & 2.43\% & 16s \\
		EMNH & \underline{0.4699}  & \underline{0.28\%} & 5s    & 0.4324  & 2.13\% & 9s    & 0.4866  & 2.15\% & 16s \\
		PMOCO-Aug & \textbf{0.4712 } & \textbf{0.00\%} & 3.2m  & 0.4409  & 0.20\% & 28m   & 0.4956  & 0.34\% & 1.7h \\
		MDRL-Aug & \textbf{0.4712 } & \textbf{0.00\%} & 2.6m  & 0.4408  & 0.23\% & 25m   & 0.4958  & 0.30\% & 1.7h \\
		EMNH-Aug & \textbf{0.4712 } & \textbf{0.00\%} & 2.6m   & \underline{0.4418}  & \underline{0.00\%} & 25m  & \underline{0.4973}  & \underline{0.00\%} & 1.7h \\
		\bottomrule
	\end{tabular}%
	\label{tab1}%
\end{table}%

\begin{table}[!t]
	\caption{Results on 200 random instances for MOCOPs with imbalanced objective domains.}
	\centering
    \addtolength{\tabcolsep}{-1pt}
	\begin{tabular}{l|ccc|ccc|ccc}
		\toprule
		& \multicolumn{3}{c|}{Bi-CVRP ($n$=20)} & \multicolumn{3}{c|}{Bi-CVRP ($n$=50)} & \multicolumn{3}{c}{Bi-CVRP ($n$=100)} \\
		Method & HV$\uparrow$    & Gap$\downarrow$   & Time  & HV$\uparrow$    & Gap$\downarrow$   & Time  & HV$\uparrow$    & Gap$\downarrow$   & Time \\
		\midrule
		PPLS/D-C & 0.4287  & 0.35\% & 1.6h  & 0.4007  & 2.41\% & 9.7h  & 0.3946  & 3.26\% & 38h \\
		\midrule
		DRL-MOA & 0.4287  & 0.35\% & 10s   & 0.4076  & 0.73\% & 12s   & 0.4055  & 0.59\% & 33s \\
		PMOCO & 0.4267  & 0.81\% & 7s    & 0.4036  & 1.70\% & 12s   & 0.3913  & 4.07\% & 32s \\
		MDRL  & 0.4291  & 0.26\% & 8s    & 0.4082  & 0.58\% & 13s   & 0.4056  & 0.56\% & 32s \\
		EMNH & \underline{0.4299}  & \underline{0.07\%} & 8s    & \underline{0.4098}  & \underline{0.19\%} & 12s   & \underline{0.4072}  & \underline{0.17\%} & 32s \\
		PMOCO-Aug & 0.4294  & 0.19\% & 13s   & 0.4080  & 0.63\% & 36s   & 0.3969  & 2.70\% & 2.7m \\
		MDRL-Aug & 0.4294  & 0.19\% & 11s   & 0.4092  & 0.34\% & 36s   & \underline{0.4072}  & \underline{0.17\%} & 2.8m \\
		EMNH-Aug & \textbf{0.4302 } & \textbf{0.00\%} & 11s   & \textbf{0.4106 } & \textbf{0.00\%} & 35s   & \textbf{0.4079 } & \textbf{0.00\%} & 2.8m \\
		\midrule
		& \multicolumn{3}{c|}{Bi-TSP-2 ($n$=20)} & \multicolumn{3}{c|}{Bi-TSP-2 ($n$=50)} & \multicolumn{3}{c}{Bi-TSP-2 ($n$=100)} \\
		Method & HV$\uparrow$    & Gap$\downarrow$   & Time  & HV$\uparrow$    & Gap$\downarrow$   & Time  & HV$\uparrow$    & Gap$\downarrow$   & Time \\
		\midrule
		WS-LKH & 0.6660  & 0.13\% & 11m   & \textbf{0.7390 } & \textbf{-0.07\%} & 1.8h  & \textbf{0.8055 } & \textbf{-0.74\%} & 6.1h \\
		PPLS/D-C & \underline{0.6662}  & \underline{0.10\%} & 27m   & 0.7300  & 1.15\% & 3.3h  & 0.7859  & 1.71\% & 10h \\
		\midrule
		DRL-MOA & 0.6657  & 0.18\% & 6s    & 0.7359  & 0.35\% & 8s    & 0.7965  & 0.39\% & 18s \\
		PMOCO & 0.6590  & 1.18\% & 5s    & 0.7347  & 0.51\% & 8s    & 0.7944  & 0.65\% & 17s \\
		MDRL  & \textbf{0.6669 } & \textbf{0.00\%} & 5s    & 0.7361  & 0.32\% & 9s    & 0.7965  & 0.39\% & 15s \\
		EMNH & \textbf{0.6669 } & \textbf{0.00\%} & 5s    & 0.7361  & 0.32\% & 9s    & 0.7965  & 0.39\% & 15s \\
		PMOCO-Aug & 0.6653  & 0.24\% & 13s   & 0.7375  & 0.14\% & 33s   & 0.7988  & 0.10\% & 3.5m \\
		MDRL-Aug & \textbf{0.6669 } & \textbf{0.00\%} & 8s   & \underline{0.7385}  & \underline{0.00\%} & 22s    & \underline{0.7996}  & \underline{0.00\%} & 2.8m \\
		EMNH-Aug & \textbf{0.6669 } & \textbf{0.00\%} & 8s   & \underline{0.7385}  & \underline{0.00\%} & 22s    & \underline{0.7996}  & \underline{0.00\%} & 2.8m \\
		\midrule
		& \multicolumn{3}{c|}{Tri-TSP-2 ($n$=20)} & \multicolumn{3}{c|}{Tri-TSP-2 ($n$=50)} & \multicolumn{3}{c}{Tri-TSP-2 ($n$=100)} \\
		Method & HV$\uparrow$    & Gap$\downarrow$   & Time  & HV$\uparrow$    & Gap$\downarrow$   & Time  & HV$\uparrow$    & Gap$\downarrow$   & Time \\
		\midrule
		WS-LKH & \textbf{0.5035 } & \textbf{0.00\%} & 13m   & \textbf{0.5305 } & \textbf{-0.49\%} & 2.0h  & \textbf{0.5996 } & \textbf{-1.70\%} & 6.6h \\
		PPLS/D-C & \textbf{0.5035 } & \textbf{0.00\%} & 1.4h  & 0.5045  & 4.42\% & 4.1h  & 0.5306  & 10.01\% & 15h \\
		\midrule
		DRL-MOA & 0.5019  & 0.32\% & 6s    & 0.5101  & 3.37\% & 8s    & 0.5488  & 6.92\% & 19s \\
		PMOCO & 0.5020  & 0.30\% & 5s    & 0.5176  & 1.95\% & 8s    & 0.5777  & 2.02\% & 18s \\
		MDRL  & \underline{0.5024}  & \underline{0.22\%} & 5s    & 0.5183  & 1.82\% & 9s    & 0.5806  & 1.53\% & 16s \\
		EMNH & \underline{0.5024}  & \underline{0.22\%} & 5s    & 0.5205  & 1.40\% & 9s    & 0.5813  & 1.41\% & 16s \\
		PMOCO-Aug & \textbf{0.5035 } & \textbf{0.00\%} & 55s   & 0.5258  & 0.40\% & 6.1m  & 0.5862  & 0.58\% & 32m \\
		MDRL-Aug & \textbf{0.5035 } & \textbf{0.00\%} & 37s   & 0.5267  & 0.23\% & 4.2m  & 0.5886  & 0.17\% & 30m \\
		EMNH-Aug & \textbf{0.5035 } & \textbf{0.00\%} & 37s   & \underline{0.5279}  & \underline{0.00\%} & 4.2m  & \underline{0.5896}  & \underline{0.00\%} & 30m \\
		\bottomrule
	\end{tabular}%
	\label{tab2}%
\end{table}%

\textbf{Optimality gap.} The gaps of HV with respect to EMNH-Aug are reported for all methods. According to the results in Tables \ref{tab1} and \ref{tab2}, EMNH outperforms other neural heuristics without instance augmentation on all problems. When equipped with instance augmentation, EMNH-Aug further improves the solution, and performs superior to most of the baselines, while slightly inferior to WS-LKH on MOTSP with $n$=50 and $n$=100. However, as iteration-based methods, WS-LKH and MOEAs take quite long running time on MOCOPs. For the neural heuristics, DRL-MOA is less agile since it trains multiple fixed models for a priori weight vectors. For new weight vectors concerned by decision makers, EMNH, MDRL, and PMOCO are able to efficiently produce high-quality trade-off solutions, where EMNH achieves the smallest gap.

\textbf{Imbalanced objective domains.}  As demonstrated in Table \ref{tab2}, the gaps between EMNH and PMOCO or MDRL are further increased on the problems with imbalanced objective domains, i.e., Bi-CVRP, Bi-TSP-2, and Tri-TSP-2. EMNH exhibits more competitive performance on these problems, revealing the effectiveness of the proposed scaled symmetric sampling method in tackling the imbalance of objective domains. We further equip PMOCO with the same sampling method, which actually also improved the solutions for those problems, but still performed inferior to EMNH (see Appendix H).

\textbf{Generalization ability.} We test the generalization ability of the model on 200 larger-scale random instances ($n$=150/200) and 3 commonly used MOTSP benchmark instances (KroAB100/150/200) in TSPLIB \cite{rei91}. The zero-shot generalization performance of the model trained and fine-tuned both on the instances with $n$=100 is reported in Appendix I. The results suggest that EMNH has a superior generalization ability compared with MOEAs and other neural heuristics for larger problem sizes.

\textbf{Hyper-parameter study.} The results in Appendix J showed that the number of sampled weight vectors $\tilde{N}=M$ is a more desirable setting and WS is a simple yet effective scalarization method.

\subsection{Learning efficiency}

As verified above, EMNH is able to produce superior solutions to the state-of-the-art neural heuristics, especially demonstrating a significant advantage over PMOCO in terms of solution quality. Then we further show that EMNH also has favorable learning efficiency against MDRL.

\textbf{Training efficiency.} EMNH, MDRL, and PMOCO only train one model to tackle all subproblems, where EMNH and MDRL use the same amount of training instances, and PMOCO requires a few more training instances according to its original setting. DRL-MOA needs to train multiple submodels with more training instances. More details are presented in Appendix F. Figure \ref{fig4a} displays the speed-up ratio, i.e., the ratio of the training time of MDRL to EMNH. Due to the multi-task based training for (partial) architecture reuse, training time of EMNH is only about $1/\tilde{N}$ of that of MDRL.

\begin{figure}[!t]
	\centering
	\subfigure[]{
		\centering
		\includegraphics[width=0.27\textwidth]{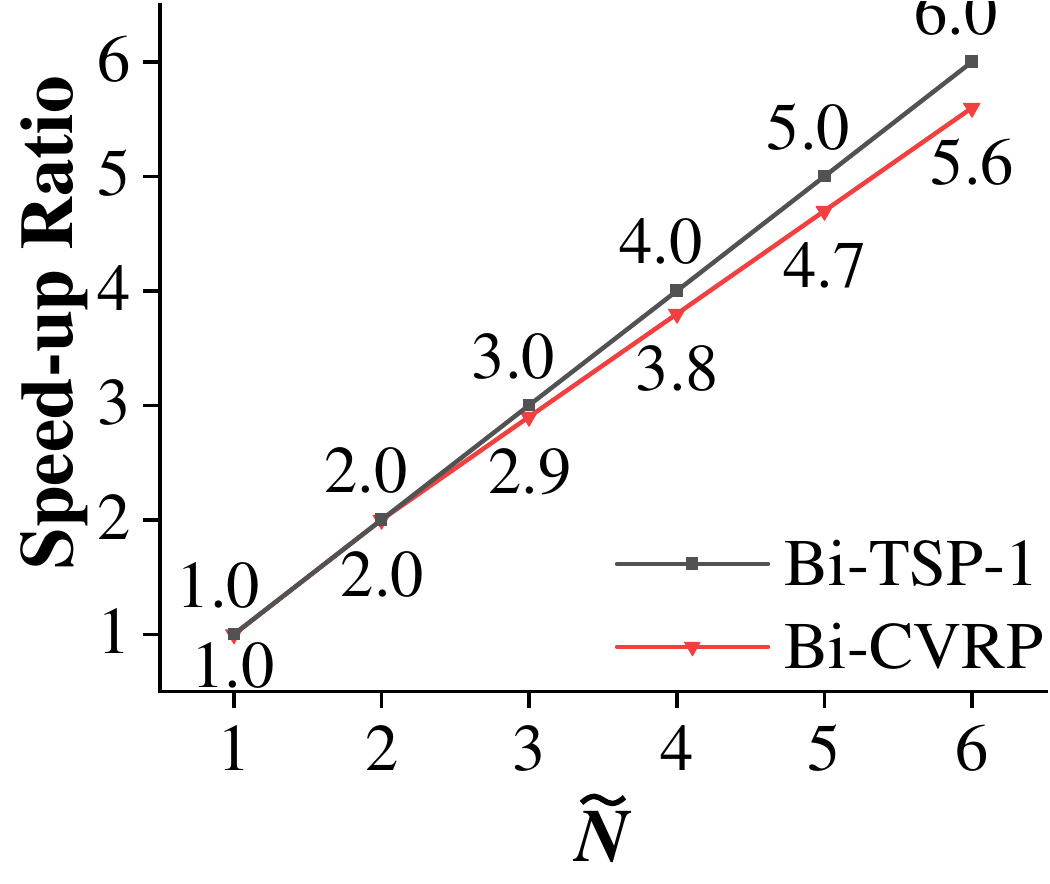}
		\label{fig4a}
	}
	\subfigure[]{
		\centering
		\includegraphics[width=0.41\textwidth]{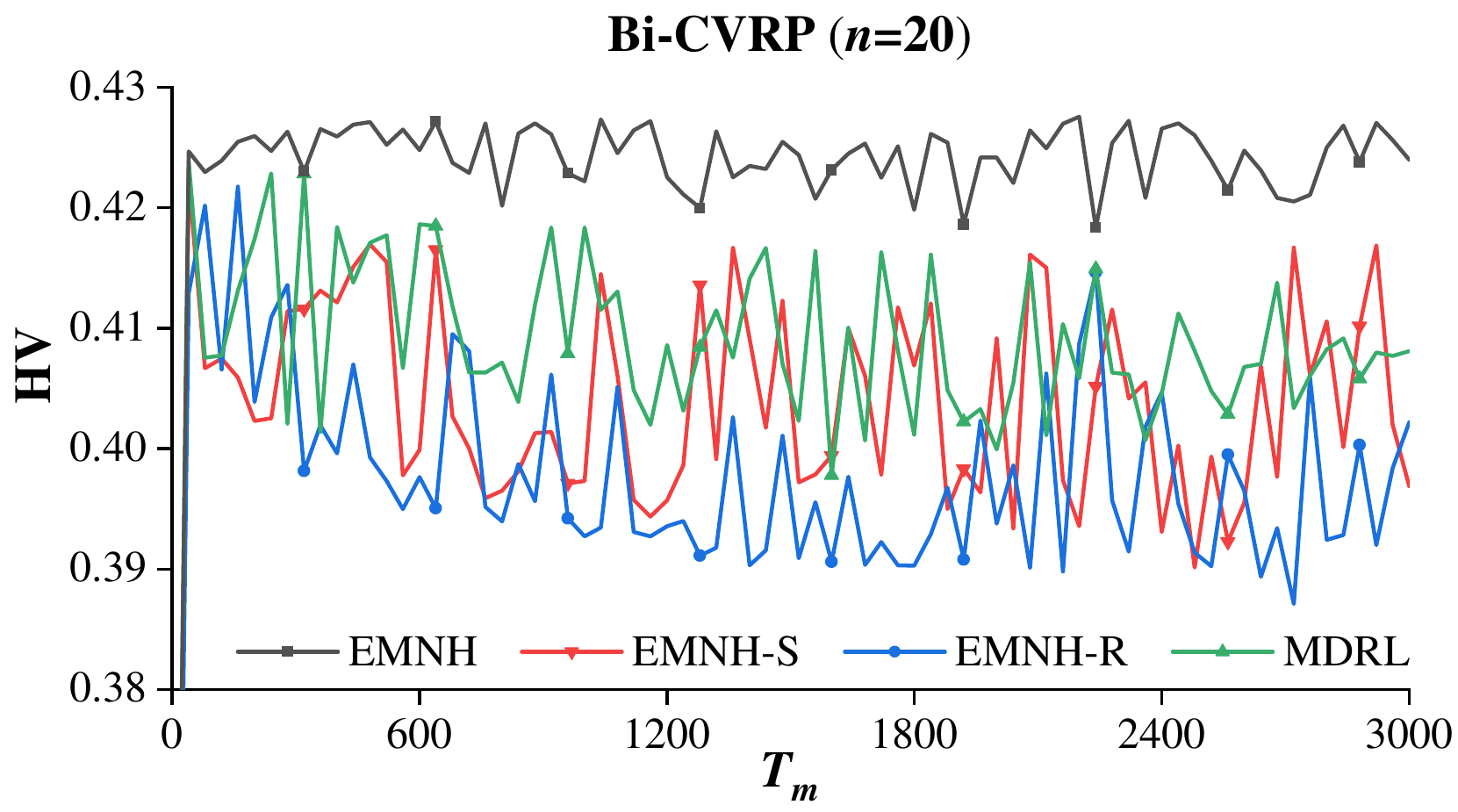}
		\label{fig4b}
	}
	\subfigure[]{
		\centering
		\includegraphics[width=0.27\textwidth]{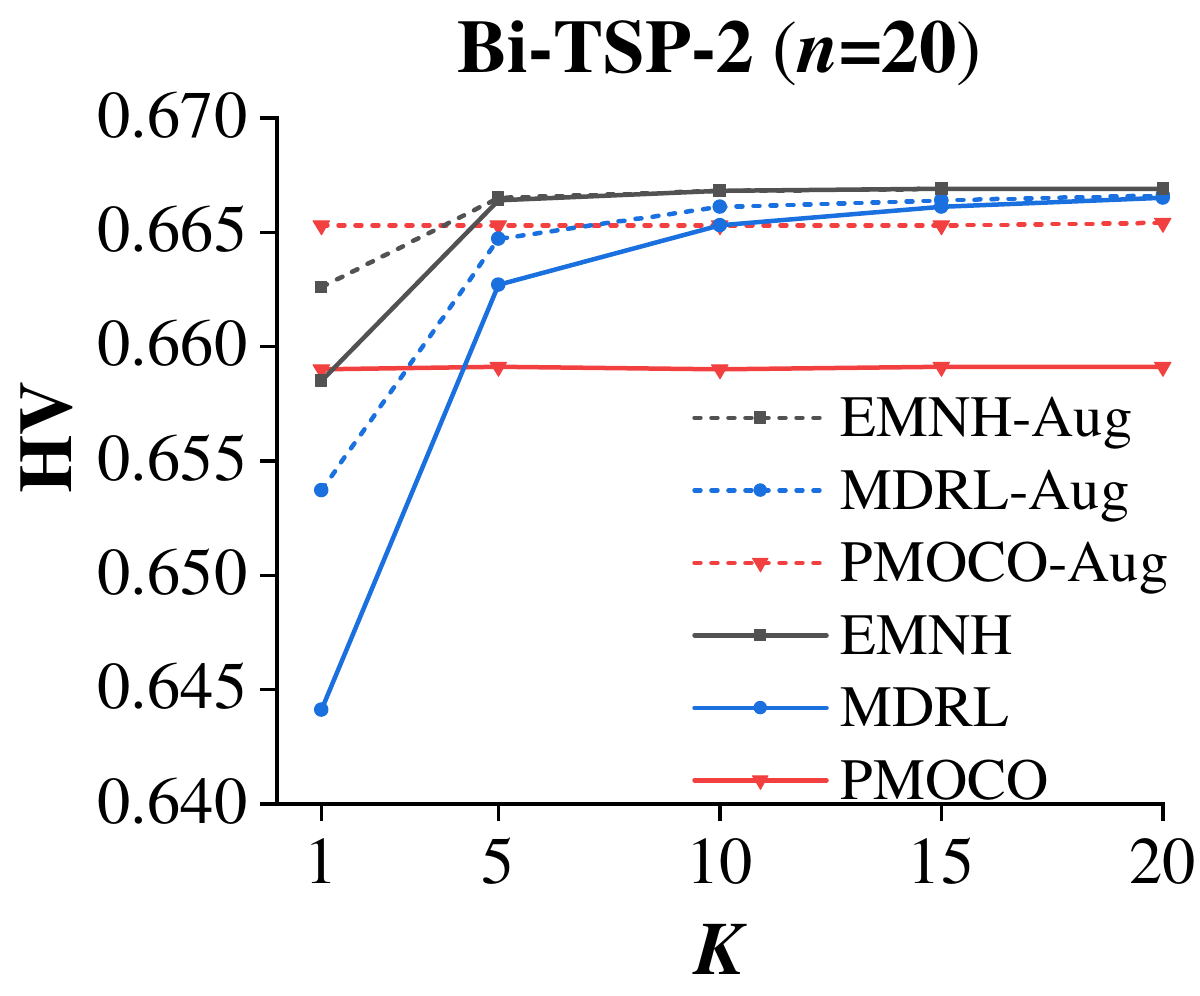}
		\label{fig4c}
	}
	\caption{Learning efficiency. (a) Training efficiency. (b) Training stability. (c) Fine-tuning efficiency.}
	\label{fig4}
\end{figure}

\textbf{Training stability.} Figure \ref{fig4b} shows the stability of the training process, where EMNH-S and EMNH-R refer to EMNH with the symmetric sampling and the random sampling, respectively. EMNH achieves the stablest and best training performance, compared to EMNH-S, EMNH-R, and MDRL, as the proposed sampling method in EMNH considers symmetric weight vectors and (imbalanced) objective domains. More details are presented in Appendix F.

\textbf{Fine-tuning efficiency.} Figure \ref{fig4c} compares the hierarchical fine-tuning in EMNH with the vanilla fine-tuning method in MDRL, where the HV of EMNH with $K$ fine-tuning steps at each level and the HV of MDRL with approximately equal total fine-tuning steps are presented. The results demonstrate that EMNH attains higher fine-tuning efficiency than MDRL, i.e., larger gaps with smaller fine-tuning steps. Furthermore, PMOCO is also equipped with the hierarchical fine-tuning (for both versions) for a fair comparison. It is worth noting that our fine-tuning process is performed individually for each weight vector on dedicated fine-tuning instances. As can be seen, PMOCO could hardly get improved by fine-tuning for zero-shot inference on test instances, since it has already converged for the corresponding subproblems. This means that EMNH has favorable potential to derive more desirable submodels to tackle specific tasks. Notably, EMNH with a few fine-tuning steps (e.g., $K=5$) outperforms PMOCO in most cases, as demonstrated in Appendix F.

\section{Conclusion}

This paper proposes an efficient meta neural heuristic (EMNH) for MOCOPs. Specifically, EMNH reduces training time, stabilizes the training process, and curtails total fine-tuning steps via (partial) architecture-shared multi-task learning, scaled symmetric sampling (of weight vectors), and hierarchical fine-tuning, respectively. The experimental results on MOTSP, MOCVRP, and MOKP verified the superiority of EMNH in learning efficiency and solution quality. A limitation is that our EMNH, as the same as other neural heuristics, can not guarantee obtaining the exact Pareto front. In the future, we will (1) extend EMNH to other MOCOPs with complex constraints; (2) investigate other advanced meta-learning algorithms and neural solvers as the base model to further improve the performance.

\section*{Acknowledgments and disclosure of funding}

This work is supported by the National Natural Science Foundation of China (62072483), and the Guangdong Basic and Applied Basic Research Foundation (2022A1515011690, 2021A1515012298).

{
\small
\bibliographystyle{unsrtnat}
\bibliography{ref}
}

\clearpage

\appendix
\setcounter{page}{1} 
\appendix

\vbox{
\hrule height 4pt
\vskip 0.25in
\vskip -\parskip%
\centering
{\LARGE\bf Efficient Meta Neural Heuristic for Multi-Objective Combinatorial Optimization (Appendix)}
\vskip 0.29in
\vskip -\parskip
\hrule height 1pt
}

    \section{Model architecture}
 
    The architecture of the base model in meta-learning is the same as POMO \cite{kwo20}, composed of an encoder and a decoder (see Figure \ref{figA1a}). For node features $\bm{x}_1, \dots, \bm{x}_n$, the encoder first computes initial node embeddings $\bm{h}^0_1, \dots, \bm{h}^0_n \in \mathcal{R}^d$ ($d=128$) by a linear projection (LP). The final node embeddings $\bm{h}^{\mathcal{N}}_1, \dots, \bm{h}^{\mathcal{N}}_n$ are further computed by $\mathcal{N}=6$ attention layers. Each attention layer is composed of a multi-head attention (MHA) with $M=8$ heads and fully connected feed-forward (FF) sublayer. Each sublayer adds a skip-connection (ADD) and batch normalization (BN).

	The decoder sequentially chooses a node according to a probability distribution produced by the node embeddings to construct a solution. The total decoding step $T$ is determined by the specific problem. At step $t$ in the decoding procedure, the \emph{query} $\bm{q}_c \in \mathcal{R}^d$ is computed by an MHA layer using the node embeddings and problem-specific \emph{context} embedding $\bm{h}_c$ (see Appendix D). The \emph{key} $\bm{k}_1, \dots, \bm{k}_n \in \mathcal{R}^d$ is computed by $\bm{k}_{i'} = W^K \bm{h}_{i'}$ for node $i'$. Then, in the final single-head attention layer, the \emph{compatibility} $\bm{u}$ is computed by $\bm{q}_c$ and $\bm{k}_{i'}$ as follows,
	\begin{equation}
		u_{i'}=\left\{
		\begin{array}{lcl}
			-\infty, & & {\rm{node}}~i'~{\rm{is~masked}}\\
			C \cdot \tanh (\frac{\bm{q}_c^T \bm{k}_{i'}}{\sqrt{d/M}}), & & \rm{otherwise}
		\end{array} \right.
	\end{equation}    
	where $C=10$ is adopted to clip the result. Finally, the probability distribution $P_{\bm{\theta}}(\bm{\pi}|s)$ to select node $i'$ is computed by the softmax function,
	\begin{equation}
		P_{i'}=P_{\bm{\theta}}(\pi_t=i'|\bm{\pi}_{1:t-1},s)=\frac{e^{u_{i'}}}{\sum_{j=1}^{n}e^{u_j}}.
	\end{equation}
	For the meta-model $\bm{\theta}$, $\bm{\theta}_{\rm{head}}$ can be defined as $W^K$ for the final single-head attention layer, and $\bm{\theta}_{\rm{body}}$ is composed of the whole encoder $\bm{\theta}_{\rm{en}}$ and the decoder body $\bm{\theta}_{\rm{de-body}}$.
	
	In each meta-iteration, we adopt a multi-task model $\bm{\tilde{\theta}}$, as shown in Figure \ref{figA1b}, to learn $\tilde{N}$ sampled tasks in parallel. The multi-task model $\bm{\tilde{\theta}}$ consists of $\bm{\tilde{\theta}}_{\rm{body}}$ and $\bm{\tilde{\theta}}_{{\rm{head}}_1}, \dots, \bm{\tilde{\theta}}_{{\rm{head}}_{\tilde{N}}}$, where $\bm{\tilde{\theta}}_{\rm{body}}$ and $\bm{\tilde{\theta}}_{{\rm{head}}_i}$ have the same architecture as $\bm{\theta}_{\rm{body}}$ and $\bm{\theta}_{\rm{head}}$, respectively. $\bm{\tilde{\theta}}_{{\rm{head}}_i}$, i.e., $W^K_{i}$, is individually updated for subproblem $i$, while $\bm{\tilde{\theta}}_{\rm{body}}$ is shared across $\tilde{N}$ tasks. Specifically, the shared node embeddings are first computed by $\bm{\tilde{\theta}}_{\rm{en}}$. Then, at step $t$ in the decoding procedure, for subproblem $i$, the \emph{query} $\bm{q}_{c,i}$ is computed using the node embeddings and \emph{context} embedding $\bm{h}_{c,i}$. The \emph{key} $\bm{k}_{1,i}, \dots, \bm{k}_{n,i}$ is computed by $\bm{k}_{i',i} = W^K_{i} \bm{h}_{i'}$. The \emph{compatibility} $\bm{u}_{i}$ is computed as follows,
	\begin{equation}
		u_{i',i}=\left\{
		\begin{array}{lcl}
			-\infty, & & {\rm{node}}~i'~{\rm{is~masked}}\\
			C \cdot \tanh (\frac{\bm{q}_{c,i}^T \bm{k}_{i',i}}{\sqrt{d/M}}), & & \rm{otherwise}
		\end{array} \right.
	\end{equation}
	Finally, the probability $P_{\bm{\tilde{\theta}}i}(\bm{\pi}|s)$ for subproblem $i$ to select node $i'$ is computed as follows,
	\begin{equation}
		P_{i',i}=P_{\bm{\tilde{\theta}}i}(\pi_t=i'|\bm{\pi}_{1:t-1},s)=\frac{e^{u_{i',i}}}{\sum_{j=1}^{n}e^{u_{j,i}}}.
	\end{equation}
	
	\section{Scaled symmetric sampling method}
	
	The scaled symmetric sampling method is shown in Algorithm \ref{algA1}. The scaled factor $f'_m$ is first estimated (Lines 1 -- 3). Then $\lfloor \tilde{N}/M \rfloor$ weight vectors are randomly sampled (Lines 5 -- 6). For each of them, $M-1$ scaled symmetric weight vectors are generated (Lines 7 -- 16).
	
	\begin{algorithm}[!t]
		\renewcommand{\algorithmicrequire}{\textbf{Input:}}
		\renewcommand{\algorithmicensure}{\textbf{Output:}}
		\caption{Scaled symmetric sampling method}
		\label{algA1}
		\begin{algorithmic}[1]
			\REQUIRE meta-model \bm{$\theta$}, problem size $n$, weight vector distribution $\Lambda$, number of objectives $M$, number of symmetric sampled weight vectors $\tilde{N}$, validation dataset $\mathcal{V}$
			\STATE $\{\bm{\pi}^k|\mathcal{V}_j\} \sim \textbf{GreedyRollout}(P_{\bm{\theta}}(\cdot|\mathcal{V}_j))$, $\quad \forall j\in\{1, \cdots, |\mathcal{V}|\}$, $\forall k \in \{1, \cdots, n\}$
			\STATE $\bm{\pi}_j \leftarrow {\rm{argmax}}_k~g(\bm{\pi}^k|(\mathcal{V}_j,\bm{1}/M))$
			\STATE $f'_m \leftarrow \frac{1}{|\mathcal{V}|}\sum_{j=1}^{|\mathcal{V}|} f_m(\bm{\pi}_j)$, $\quad \forall m \in \{1, \cdots, M\}$
			\FOR {$i = 1$ to $\tilde{N}$}
			\IF {$i \leq \lfloor \tilde{N}/M \rfloor$}
			\STATE $\bm{\lambda}_i \sim \textbf{SampleWeight}(\Lambda)$
			\ELSIF {$\lfloor \tilde{N}/M \rfloor < i \leq M \times \lfloor \tilde{N}/M \rfloor$}
			\FOR {$m = 1$ to $M$}
			\STATE $\bm{\lambda}'_i \leftarrow \bm{\lambda}_i$
			\IF {$m=1$}
			\STATE $\lambda_{i,m} \leftarrow \lambda'_{i-\lfloor \tilde{N}/M \rfloor,M}\times f'_M/f'_m$
			\ELSE
			\STATE $\lambda_{i,m} \leftarrow \lambda'_{i-\lfloor \tilde{N}/M \rfloor,m-1}\times f'_{m-1}/f'_m$
			\ENDIF
			\ENDFOR
			\STATE $\bm{\lambda}_i \leftarrow \bm{\lambda}_i/\sum_{m=1}^{M}\lambda_{i,m}$
			\ELSE
			\STATE $\bm{\lambda}_i \sim \textbf{SampleWeight}(\Lambda)$
			\ENDIF
			\ENDFOR
			\ENSURE $\{\bm{\lambda}_1, \cdots, \bm{\lambda}_{\tilde{N}}\}$
		\end{algorithmic}  
	\end{algorithm}

	\section{Hierarchical fine-tuning method}
	
	We consider an $L$-level $a$-section hierarchy. The whole weight space is uniformly divided into $N^{(l)}$ subspaces in level $l$. $N^{(l)}$ weight vectors are the centers of these subspaces. In level $l+1$, the $N^{(l)}$ submodels are fine-tuned to derive $N^{(l+1)}=aN^{(l)}$ submodels with $K^{(l+1)}$ fine-tuning steps, which are the centers of $N^{(l+1)}$ subspaces. The $j$-th submodel in level $l+1$ is fine-tuned from the $i$-th submodel in level $l$, where the $j$-th weight vector belongs to the $i$-th subspace in level $l$.
	
\begin{figure}[!t]
		\centering
		\subfigure[]{
			\centering
			\includegraphics[width=0.95\textwidth]{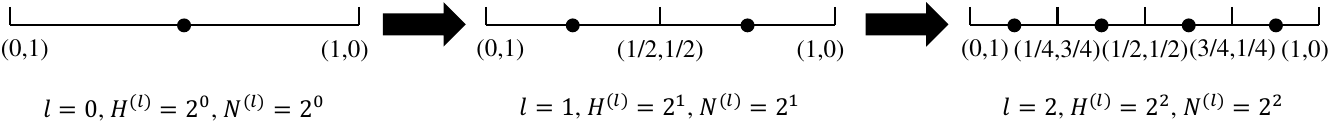}
			\label{figA2a}
		}
		\subfigure[]{
			\centering
			\includegraphics[width=0.95\textwidth]{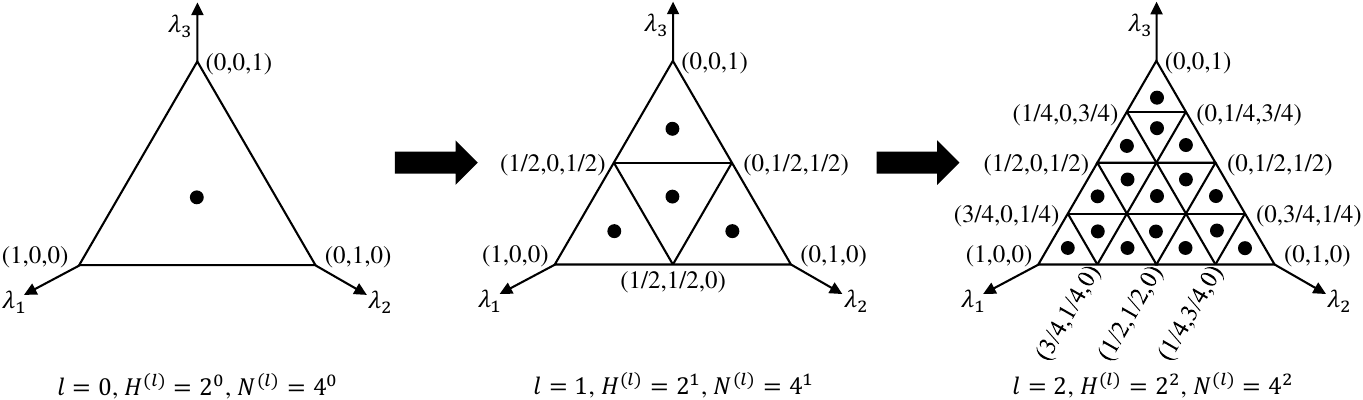}
			\label{figA2b}
		}
		\caption{The hierarchy of fine-tuning. (a) $M=2$. (b) $M=3$.}
		\label{figA2}
	\end{figure}

	The uniform division of the weight space is illustrated as follows. According to the \citet{das98} method, $C^{M-1}_{H+M-1}$ vertices can be generated, where $M$ is the number of objectives and $H$ is a user-defined hyper-parameter. Then, the whole weight space is uniformly divided by $C^{M-1}_{H+M-1}$ vertices. The division with $H^{(l)}=2^l$ can be seen in Figure \ref{figA2}. There are $2^l$ subspaces for $M=2$ and $4^l$ subspaces for $M=3$ in level $l$.
	
	Note that the final $N$ weights are given beforehand. Thus, we set $N^{(L)}=N$, where $N^{(L-1)}=a^{L-1}<N$ and $a^L\geq N$, i.e., $L=7$ for $M=2$ and $L=4$ for $M=3$. The final $N$ submodels are fine-tuned by $N^{(L-1)}$ submodels.
	
	For the meta-model or a coarse-tuned submodel in level $l$ ($l<L$), its fine-tuning process with a given weight vector and $K$ fine-tuning steps is shown in Algorithm \ref{algA2}.
	
	\begin{algorithm}[!t]
		\renewcommand{\algorithmicrequire}{\textbf{Input:}}
		\renewcommand{\algorithmicensure}{\textbf{Output:}}
		\caption{Fine-tuning process in each level}
		\label{algA2}
		\begin{algorithmic}[1]
			\REQUIRE the meta-model or a coarse-tuned submodel \bm{$\theta$}, a given weight vector $\bm{\lambda}$, fine-tuning steps $K$, batch size $B$, problem size $n$
			\FOR {$k = 1$ to $K$}
			\STATE $s_j \sim \textbf{SampleInstance}(\mathcal{S})$, $\quad \forall j \in \{1, \dots, B\}$
			\STATE $\{\bm{\pi}^k|s_j,\bm{\lambda}\} \sim \textbf{SampleRollout}(P_{\bm{\tilde{\theta}}i}(\cdot|s_j))$, $\quad \forall j \in \{1, \dots, B\}$, $\forall k \in \{1, \dots, n\}$
			\STATE $b_{j} \leftarrow \frac{1}{n}\sum_{k=1}^n g(\bm{\pi}^k|s_j,\bm{\lambda})$
			\STATE $\nabla\mathcal{L}(\bm{\tilde{\theta}}) \leftarrow \frac{1}{Bn}\sum\limits_{j=1}^B\sum\limits_{k=1}^n[(g(\bm{\pi}^k|s_j,\bm{\lambda})-b_{j})\nabla\text{log}P_{\bm{\theta}}(\bm{\pi}^k|s_j)]$
			\STATE $\bm{\theta} \leftarrow \textbf{Adam}(\bm{\theta}, \nabla\mathcal{L}(\bm{\theta}))$
			\ENDFOR
			\ENSURE The fine-tuned submodel $\bm{\theta}$
		\end{algorithmic}  
	\end{algorithm}
	
	\section{Details of MOCOPs}
	
	\subsection{MOTSP}
	
	\subsubsection{Problem definition}
	
	The multi-objective traveling salesman problem (MOTSP) is defined on a complete graph with $n$ nodes and $M$ cost matrices. Node $i$ has $M$ 2D coordinates $\{\bm{x}^1_i, \dots, \bm{x}^M_i\}$, where the $m$-th cost between node $i$ and $j$ is given by the Euclidean distance $c^m_{ij} = \Vert \bm{x}^m_i - \bm{x}^m_j \Vert_2$. The goal is to find a permutation $\bm{\pi}$ to minimize all the $M$ costs simultaneously, as follows,
	\begin{equation}
		\min\; \bm{f}(\bm{\pi}) = (f_1(\bm{\pi}), f_2(\bm{\pi}), \dots, f_M(\bm{\pi})),
	\end{equation}
	\begin{equation}
		{\rm{where}}\; f_m(\bm{\pi}) = c^m_{\pi_n,\pi_1}+\sum_{j=1}^{n-1}c^m_{\pi_j,\pi_{j+1}}.
	\end{equation}
	For the $M$-objective TSP type 2, each node has $(M-1)$ 2D coordinates and one number interpreted as its altitude. The single-objective counterpart, TSP, is a well-known NP-hard problem. It appears that MOTSP is even harder. Thus, its approximate Pareto optimal solutions are commonly pursued.
	
	\subsubsection{Instance}
	
	For the $M$-objective TSP type 1, each node has $M$ 2D coordinates. The random instances with $n$ nodes are sampled from uniform distribution on $[0,1]^{2M}$. For the $M$-objective TSP type 2, each node has $(M-1)$ 2D coordinates and one number. The random instances with $n$ nodes are sampled from uniform distribution on $[0,1]^{2M-1}$.
	
	\subsubsection{Model details}
	
	The input dimension of the $M$-objective TSP type 1 is $2M$ for the encoder. The input dimension of the $M$-objective TSP type 2 is $2M-1$ for the encoder.
	
	For MOTSP, POMO \cite{kwo20} uses $n$ context embedding $\bm{h}^1_c,\dots,\bm{h}^n_c$ to calculate the probability of node selection in the decoder. At the decoding step $t$, $\bm{h}^k_c$ is defined as follows,
	\begin{equation}
		\bm{h}^k_c=\left\{
		\begin{array}{lcl}
			[\bar{\bm{h}}^k;\bm{h}^k_{\bm{\pi}_{t-1}};\bm{h}^k_{\bm{\pi}_1}], & & t>1\\
			\rm{none}, & & t=1
		\end{array} \right.
	\end{equation}
	where $[;]$ is the concatenation and the graph embedding $\bar{\bm{h}}^k=\sum_{i=1}^{n} \bm{h}^k_i$. For $t=1$, the first node is not selected by the decoder. Instead, it is defined as $\bm{h}^k_{\bm{\pi}_1}=\bm{h}_k$. In the decoding procedure, the nodes already visited need to be masked.
	
	\subsection{MOCVRP}
	
	\subsubsection{Problem definition}
	
	The capacitated vehicle routing problem (CVRP), a classical extension of TSP, contains $n$ customer nodes and a depot node. Each node has a 2D coordinate. In addition, customer $i$ has a demand $d_i$ to be satisfied. A fleet of homogeneous vehicles with identical capacity $D$ is initially placed at the depot. Vehicles must serve all the customers and finally return to the depot. The capacity constraints must be satisfied, i.e., the remaining capacity of vehicles for serving customer $i$ must be no less than $d_i$.
	
	For the multi-objective capacitated vehicle routing problem (MOCVRP), we consider two conflicting objectives, i.e., the total traveling distance and the traveling distance of the longest route (makespan).
	
	\subsubsection{Instance}
	
	For MOCVRP, the coordinates of the depot and customers are sampled from uniform distribution on $[0,1]^2$. Following the previous work \cite{kwo20,lin22}, the demand $d_i$ is randomly chosen from $\{1,\dots,9\}$ and the capacity is set to $D=30/40/50$ for $n=20/50/100$. Without loss of generality, the demand and capacity are normalized as $\hat{d}_i=d_i/D$ and $\hat{D}=D/D=1$, respectively.
	
	\subsubsection{Model details}
	
	The inputs of MOCVRP are a 2D vector of the depot and $n$ 3D vectors of the customers for the encoder. Their embeddings with both 128 dimensions are obtained by two linear projections with two separate parameter matrices.
	
	For MOCVRP, the context embedding in the decoder is defined as the concatenation of the graph embedding, the embedding of the last node, and the remaining capacity. Since the first node of CVRP must be the depot node, POMO assigns the second node, i.e., the first customer node, to produce $n$ various solutions. In the decoding procedure, the nodes already visited and the nodes with a larger demand than the remaining capacity need to be masked.
	
	\subsection{MOKP}
	
	\subsubsection{Problem definition}
	
	Knapsack problem (KP) is also a representative combinatorial optimization problem. The multi-objective 0-1 knapsack problem (MOKP) with $M$ objectives and $n$ items is defined as follows,
	\begin{equation}
		\max\; \bm{f}(\bm{x}) = (f_1(\bm{x}), f_2(\bm{x}), \dots, f_M(\bm{x})),
	\end{equation}
	\begin{equation}
		{\rm{where}}\; f_m(\bm{x}) = \sum_{i=1}^{n} v_i^m x_i,
	\end{equation}
	\begin{equation}
		{\rm{subject~to}}\; \sum_{i=1}^{n} w_i x_i \leq W,
	\end{equation}
	\begin{equation}
		x_i \in \{0,1\},
	\end{equation}
	where item $i$ has a weight $w_i$ and $M$ different values \{$v_i^1,v_i^2,\dots,v_i^M$\}. $W$ is the weight capacity of the knapsack. The goal is to find a solution $\bm{x}$ to maximize all the $M$ total values simultaneously. The single-objective KP is also an NP-hard problem, so MOKP is even harder.
	
	\subsubsection{Instance}
	
	As in the previous work \cite{kwo20,lin22}, the values and weight for each item of MOKP are all sampled from uniform distribution on $[0,1]$. The knapsack capacity is set to $W=12.5/25/25$ for $n=50/100/200$.
	
	\subsubsection{Model details}
	
	For the $M$-objective MOKP, as each item has $M$ values and one weight, the input dimension is $M+1$ for the encoder. The policy network of MOKP is as the same as that of MOTSP. In the decoding procedure, the context embedding is defined as the concatenation of the graph embedding and the remaining capacity. The items already selected and the items with a larger weight than the remaining capacity need to be masked.
	
	\section{Hypervolume indicator}
	
	The hypervolume (HV) indicator is widely used to evaluate the performance of the methods for MOCOPs, since HV can comprehensively measure the convergence and diversity of $\mathcal{PF}$ without the ground truth Pareto front \cite{aud21}. For a given $M$-objective $\mathcal{PF}$ and a reference point $\bm{r}^*$, ${\rm{HV}}(\mathcal{PF}, \bm{r}^*)$ is defined as follows,
	\begin{equation}
		{\rm{HV}}(\mathcal{PF}, \bm{r}^*)=\mu(S),
	\end{equation}
	\begin{equation}
		S = \{\bm{r}\in\mathcal{R}^M | \exists \bm{r} \in \mathcal{PF}~{\rm{such~that}}~\bm{y} \prec \bm{r} \prec \bm{r}^*\},
	\end{equation}
	where $r^*_i>\max\{f_i(x)|\bm{f}(x) \in \mathcal{PF}\}$ (or $r^*_i<\min\{f_i(x)|\bm{f}(x) \in \mathcal{PF}\}$ for the maximization problem), $\forall i \in \{1,\dots, M\}$, and $\mu$ is the Lebesgue measure. An example with $M=2$ is presented in Figure \ref{figA3}, where $\mathcal{PF}=\{\bm{f}(x_1),\bm{f}(x_2),\bm{f}(x_3),\bm{f}(x_4)\}$. ${\rm{HV}}(\mathcal{PF}, \bm{r}^*)$ is equal to the size of the grey area in this example.
	
	Since HV value would considerably vary with the objective domain, we record the normalized HV ratio ${\rm{HV}}'(\mathcal{PF}, \bm{r}^*)={\rm{HV}}(\mathcal{PF}, \bm{r}^*)/\prod_{i=1}^{M}|r^*_i-z^*_i|$, where $\bm{z}^*$ is an ideal point satisfying $z^*_i<\min\{f_i(x)|\bm{f}(x) \in \mathcal{PF}\}$ (or $z^*_i>\max\{f_i(x)|\bm{f}(x) \in \mathcal{PF}\}$ for the maximization problem), $\forall i \in \{1,\dots, M\}$. For a problem, all methods share the same $\bm{r}^*$ and $\bm{z}^*$, as shown in Table \ref{tab:ref}.

    \begin{table}[!t]
	\begin{minipage}{0.54\textwidth}
	\centering
		\includegraphics[width=0.95\textwidth]{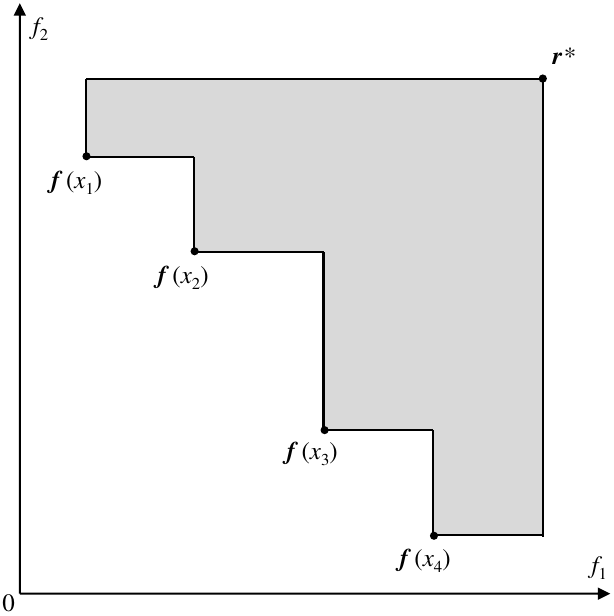} 
		\caption{Hypervolume illustration.}
		\label{figA3}
	\end{minipage}
	\hfill
	\begin{minipage}{0.44\textwidth}
		\centering
      \caption{Reference points and ideal points.}
        \resizebox{\textwidth}{!}{
        \begin{tabular}{cccc}
        \toprule
        Problem & Size  & $\bm{r}^*$     & $\bm{z}^*$ \\
        \midrule
        \multirow{5}[2]{*}{Bi-TSP-1} & 20    & (20, 20) & (0, 0) \\
              & 50    & (35, 35) & (0, 0) \\
              & 100   & (65, 65) & (0, 0) \\
              & 150   & (85, 85) & (0, 0) \\
              & 200   & (115, 115) & (0, 0) \\
        \midrule
        \multirow{3}[2]{*}{Bi-KP} & 50    & (5, 5) & (30, 30) \\
              & 100   & (20, 20) & (50, 50) \\
              & 200   & (30, 30) & (75, 75) \\
        \midrule
        \multirow{3}[2]{*}{Tri-TSP-1} & 20    & (20, 20, 20) & (0, 0) \\
              & 50    & (35, 35, 35) & (0, 0) \\
              & 100   & (65, 65, 65) & (0, 0) \\
        \midrule
        \multirow{3}[2]{*}{Bi-CVRP} & 20    & (30, 4) & (0, 0) \\
              & 50    & (45, 4) & (0, 0) \\
              & 100   & (80, 4) & (0, 0) \\
        \midrule
        \multirow{3}[2]{*}{Bi-TSP-2} & 20    & (20, 12) & (0, 0) \\
              & 50    & (35, 25) & (0, 0) \\
              & 100   & (65, 45) & (0, 0) \\
        \midrule
        \multirow{3}[2]{*}{Tri-TSP-2} & 20    & (20, 20, 12) & (0, 0) \\
              & 50    & (35, 35, 25) & (0, 0) \\
              & 100   & (65, 65, 45) & (0, 0) \\
        \bottomrule
        \end{tabular}%
        }
      \label{tab:ref}%
	\end{minipage}
    \end{table}

	\section{More details of learning efficiency}
	
	\subsection{Training efficiency}
	
	EMNH uses almost the equal number of training instances to MDRL \cite{zha22} and PMOCO  \cite{lin22}. Attributed to the accelerated training method, EMNH only consumes approximately $1/\tilde{N}$ training time of MDRL. For EMNH, we set $T_m=3000$, $T_u=100$, and $B=64$ to train the meta-model, i.e., overall $1.92 \times 10^7$ training instances. MDRL adopts the same training hyper-parameters as EMNH, i.e., the same amount of training instances. According to the settings of PMOCO, it trains the model by 200 epochs, with $1 \times 10^5$ instances at each epoch, i.e., overall $2 \times 10^7$ training instances, a little more than that of EMNH. The total training time of PMOCO is close to that of EMNH.
	
	As a multi-model method, DRL-MOA \cite{lik21} needs to train multiple networks with more training instances to deal with multiple subproblems. DRL-MOA first trains a submodel for the first weight vector by 200 epochs with $1 \times 10^5$ instances at each epoch, and then transfers its parameter by 5 epochs to derive another submodel for its neighbor subproblem. With a sequential parameter-transfer process, a set of submodels are obtained for various pre-given weight vectors. For the bi/tri-objective problems, the number of subproblems $N$ is set to 101/105. Hence, DRL-MOA needs to train 101/105 submodels with total 700/720 training epochs.
	
	\subsection{Training stability}
	
\begin{figure}[!t]
		\centering
            \includegraphics[width=0.7\textwidth]{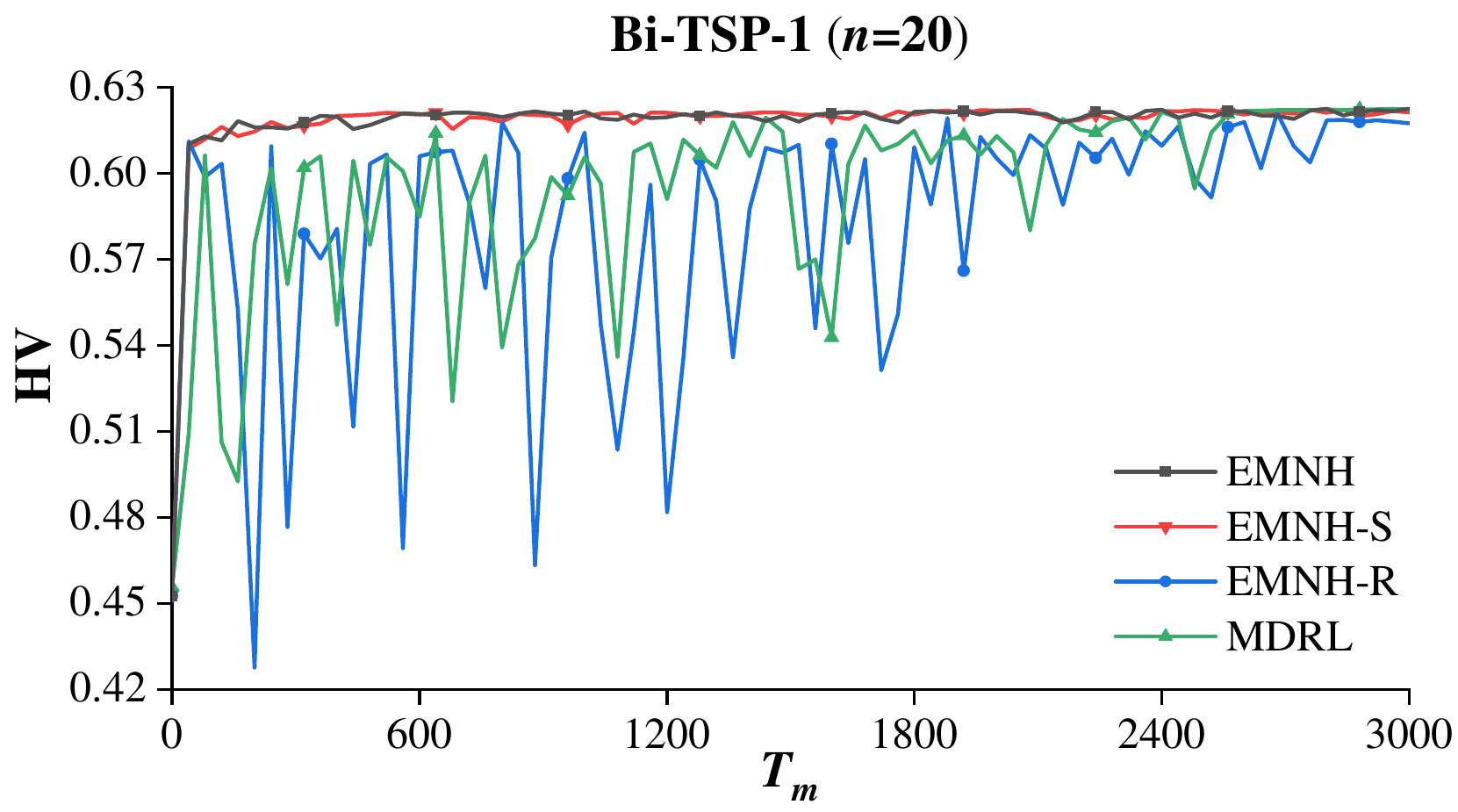}
		\caption{Training Stability on MOCOPs with balanced objective domains.}
		\label{figA4a}
	\end{figure}

	Figure \ref{figA4a} shows the results on Bi-TSP-1 with the balanced objective domains. EMNH-S and EMNH-R denote EMNH with the symmetric sampling method and the random sampling method, respectively. HV is computed by fine-tuning the meta-model with $K=1$ step at each level. The results show that EMNH achieves the stablest training process. The performance of EMNH and EMNH-S are close, while the training processes of other two methods with random sampling are unstable. However, for Bi-CVRP with the imbalanced objective domains, EMNH, which adjusts the sampled weight vectors by the objective domains, outperforms other methods, as shown in Figure \ref{fig4b}.
	
	Table \ref{tabA2} reports the final HV indicators of the trained model, which also exhibits higher solution quality of EMNH. The best result and its statistically indifferent results using a Wilcoxon rank-sum test at 1\% level are highlighted as \textbf{bold}. The second-best result and its statistically indifferent results are highlighted as \underline{underline}. The method with ``-Aug" represents the inference results using instance augmentation (see Appendix G) of POMO.

	\begin{table}[!t]
		\caption{Solution quality of the scaled symmetric sampling method.}
		\centering
		\begin{tabular}{cc|cc|cc}
			\toprule
			&       & \multicolumn{2}{c|}{$n$=20} & \multicolumn{2}{c}{$n$=50} \\
			& Method & HV$\uparrow$ & Gap$\downarrow$   & HV$\uparrow$ & Gap$\downarrow$ \\
			\midrule
			\multirow{8}*{\rotatebox{90}{Bi-TSP-1}} & MDRL  & \textbf{0.6271} & \textbf{0.00\%} & 0.6374  & 0.53\% \\
			& EMNH-R & 0.6269  & 0.04\% & 0.6339  & 1.07\% \\
			& EMNH-S & \underline{0.6270}  & \underline{0.02\%} & 0.6362  & 0.72\% \\
			& EMNH & \textbf{0.6271} & \textbf{0.00\%} & 0.6362  & 0.72\% \\
			& MDRL-Aug & \textbf{0.6271} & \textbf{0.00\%} & \textbf{0.6408} & \textbf{0.00\%} \\
			& EMNH-R-Aug & 0.6269  & 0.04\% & \underline{0.6391}  & \underline{0.26\%} \\
			& EMNH-S-Aug & \underline{0.6270}  & \underline{0.02\%} & \textbf{0.6408} & \textbf{0.00\%} \\
			& EMNH-Aug & \textbf{0.6271} & \textbf{0.00\%} & \textbf{0.6408} & \textbf{0.00\%} \\
			\midrule
			\multirow{8}*{\rotatebox{90}{Bi-CVRP}} & MDRL  & 0.4291  & 0.26\% & 0.4082  & 0.58\% \\
			& EMNH-R & 0.3774  & 12.26\% & 0.3743  & 8.83\% \\
			& EMNH-S & 0.4158  & 3.34\% & 0.3825  & 6.84\% \\
			& EMNH & \underline{0.4299}  & \underline{0.07\%} & \underline{0.4098}  & \underline{0.19\%} \\
			& MDRL-Aug & 0.4294  & 0.19\% & 0.4092  & 0.34\% \\
			& EMNH-R-Aug & 0.3849  & 10.53\% & 0.3786  & 7.78\% \\
			& EMNH-S-Aug & 0.4179  & 2.87\% & 0.3870  & 5.76\% \\
			& EMNH-Aug & \textbf{0.4302} & \textbf{0.00\%} & \textbf{0.4106} & \textbf{0.00\%} \\
			\bottomrule
		\end{tabular}%
		\label{tabA2}%
	\end{table}%
	
	\subsection{Fine-tuning efficiency}
	To study the fine-tuning efficiency, we compare the hierarchical fine-tuning method of EMNH with the vanilla fine-tuning method of MDRL. For $K$ steps at each level with $M=2$ in EMNH, the total fine-tuning step is $\sum_{l=1}^{L-1}2^lK+NK=227K$ with $N=101$, where $L$ satisfies $2^{L-1}<N$ and $2^L\geq N$, i.e., $L=7$. For MDRL, each submodel is fine-tuned directly from the meta-model with $\tilde{K}$ steps, so the total fine-tuning step is $N\tilde{K}=101\tilde{K}$. Thus, for $K=1/5/10/15/20$, $\tilde{K}=2/11/22/34/45$ ensures that the total fine-tuning step of MDRL is approximately equal to that of EMNH. For $M=3$ with $N=105$, $K$ is set to $25$ for EMNH, while $\tilde{K}$ is set to $45$ for MDRL to make the equal total of fine-tuning steps.
	
\begin{figure}[!t]
		\centering
		\subfigure[]{
			\centering
			\includegraphics[width=0.315\textwidth]{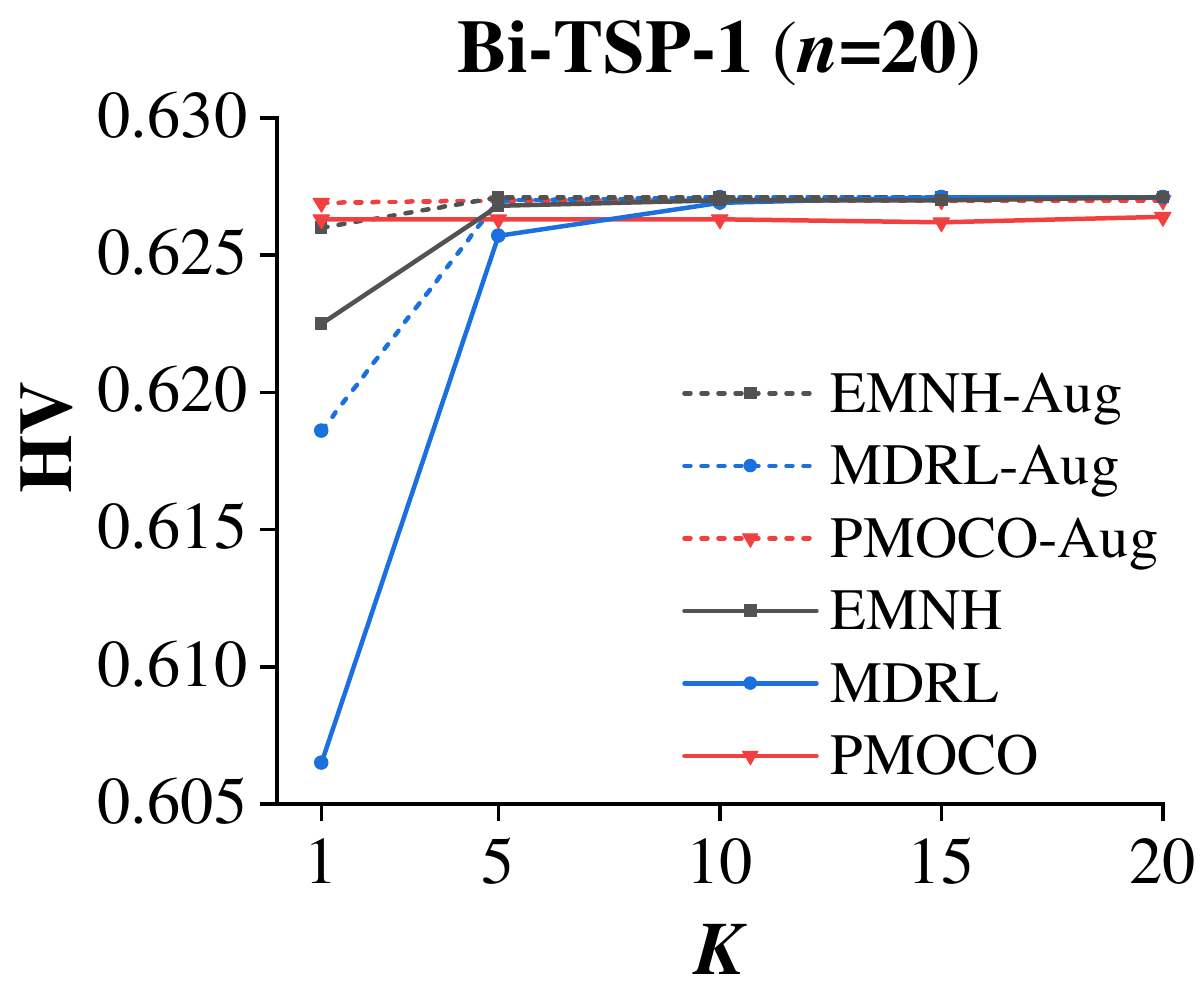}
			\label{figA5a}
		}
		\subfigure[]{
			\centering
			\includegraphics[width=0.315\textwidth]{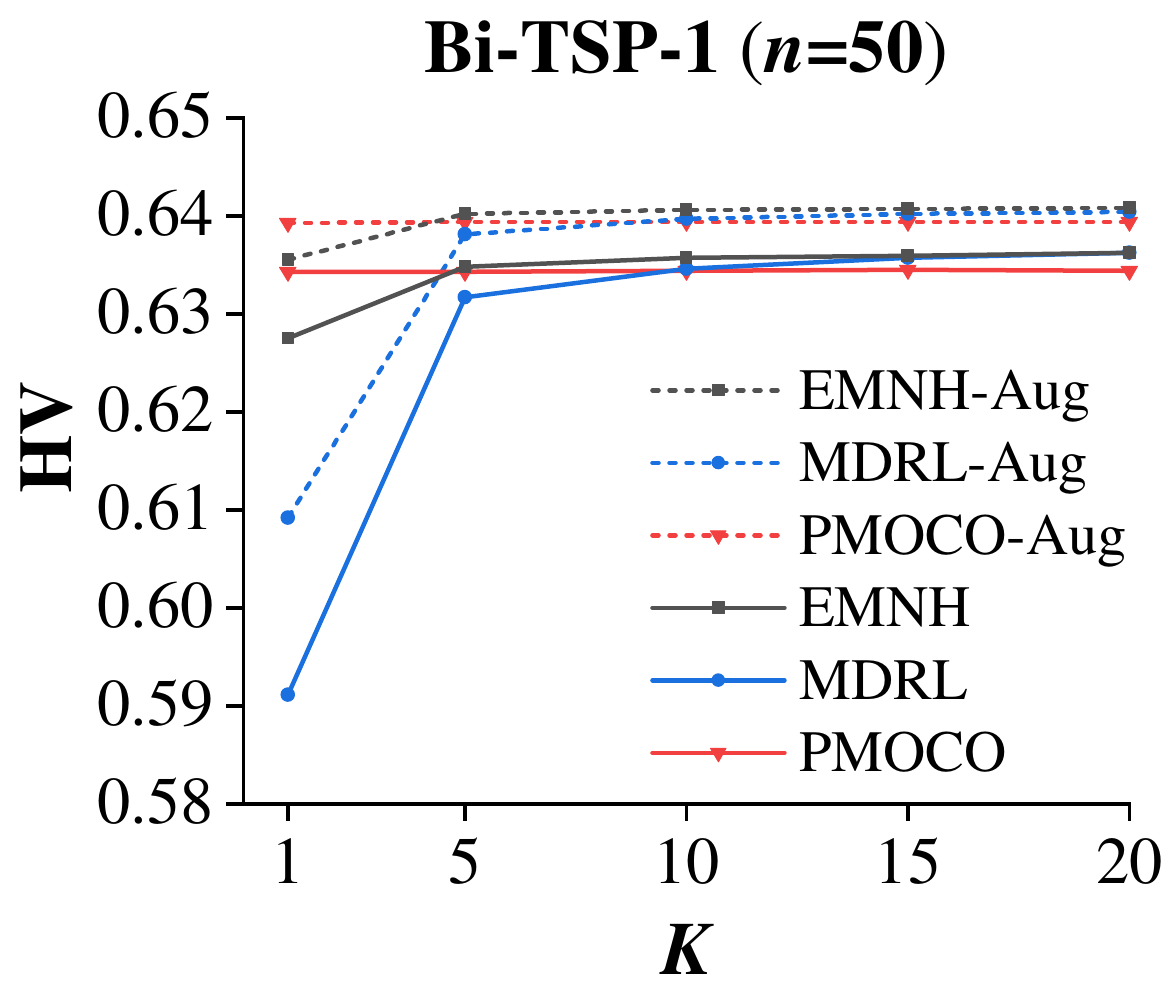}
			\label{figA5b}
		}
		\subfigure[]{
			\centering
			\includegraphics[width=0.315\textwidth]{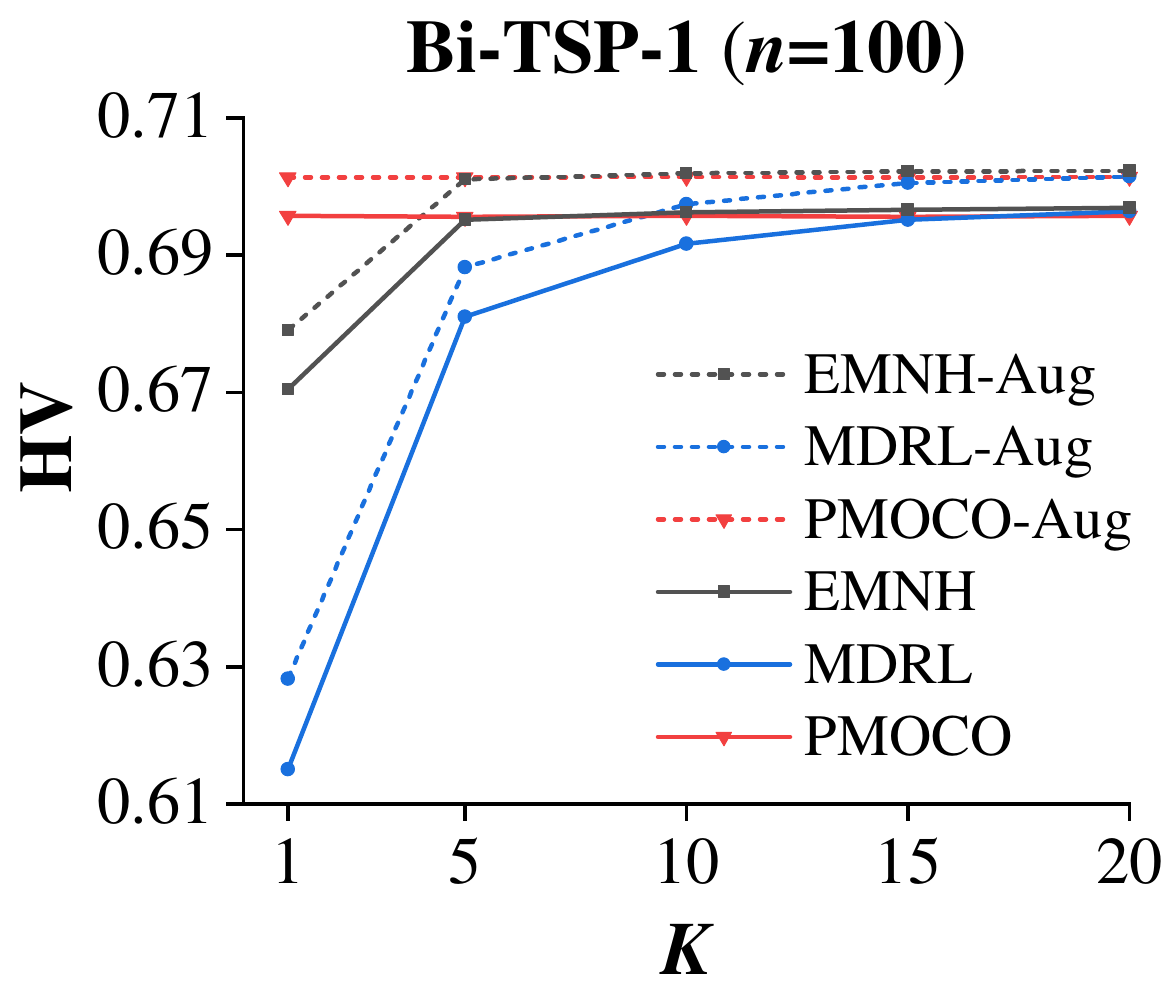}
			\label{figA5c}
		}
		\subfigure[]{
			\centering
			\includegraphics[width=0.315\textwidth]{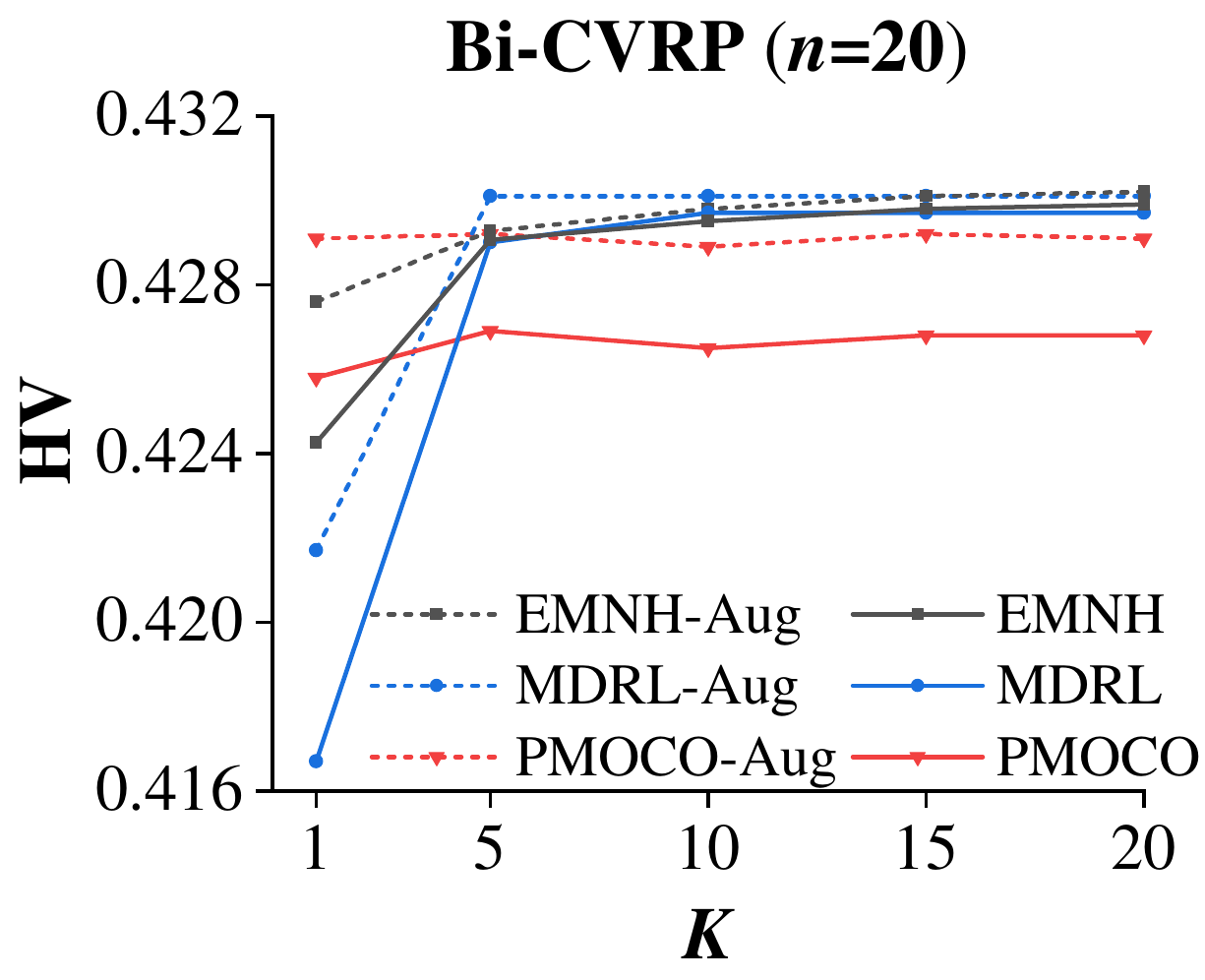}
			\label{figA5d}
		}
		\subfigure[]{
			\centering
			\includegraphics[width=0.315\textwidth]{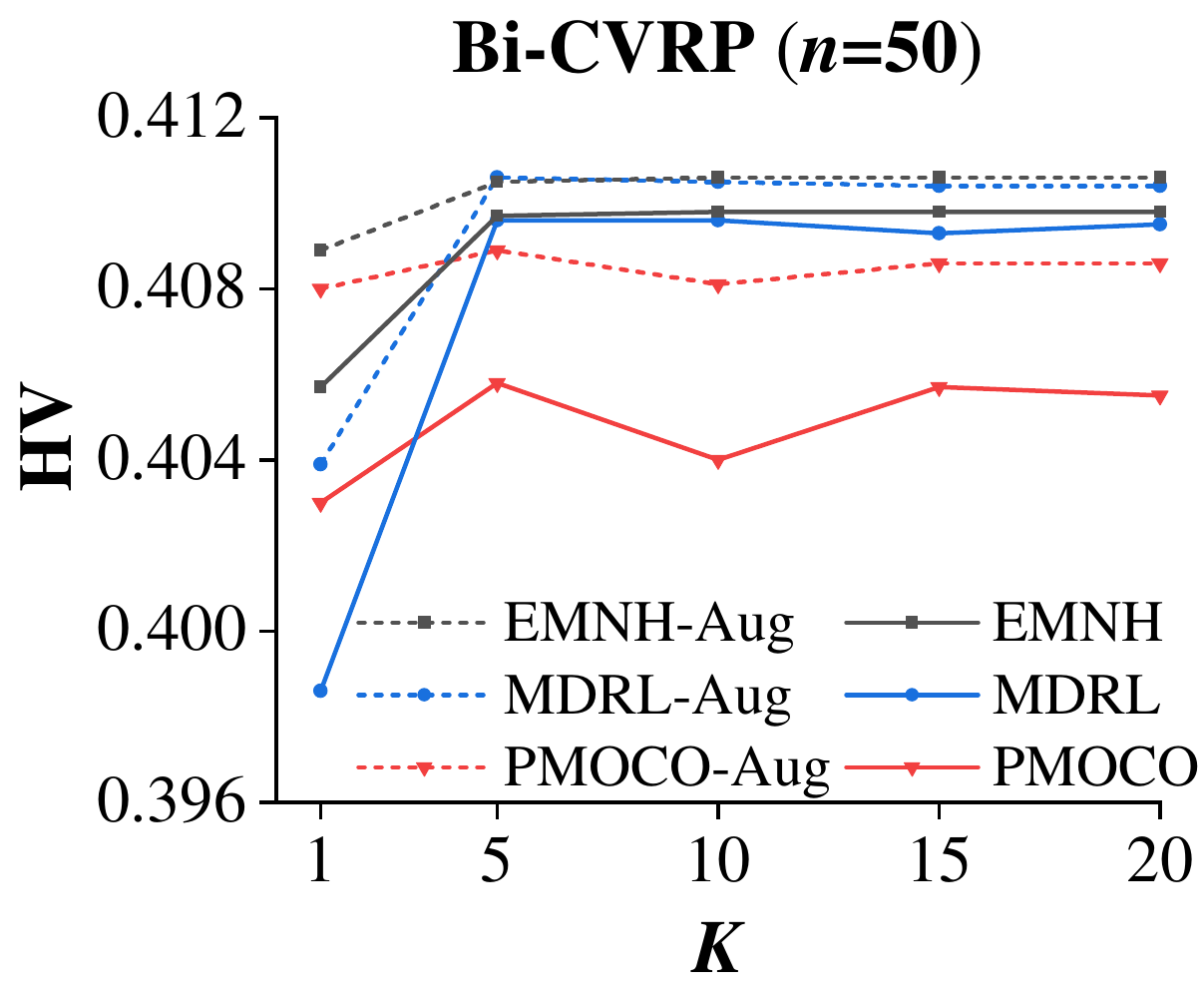}
			\label{figA5e}
		}
		\subfigure[]{
			\centering
			\includegraphics[width=0.315\textwidth]{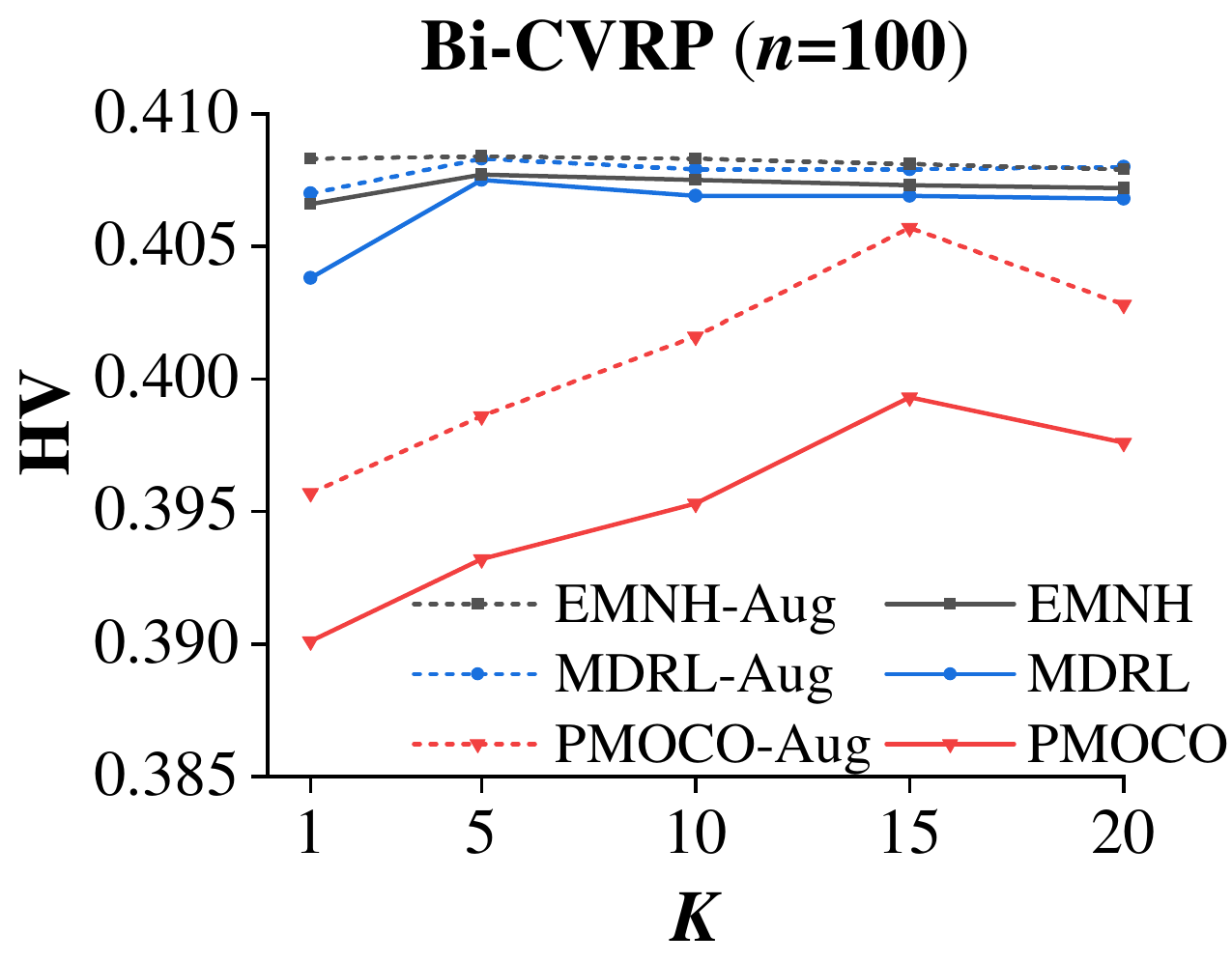}
			\label{figA5f}
		}
		\subfigure[]{
			\centering
			\includegraphics[width=0.315\textwidth]{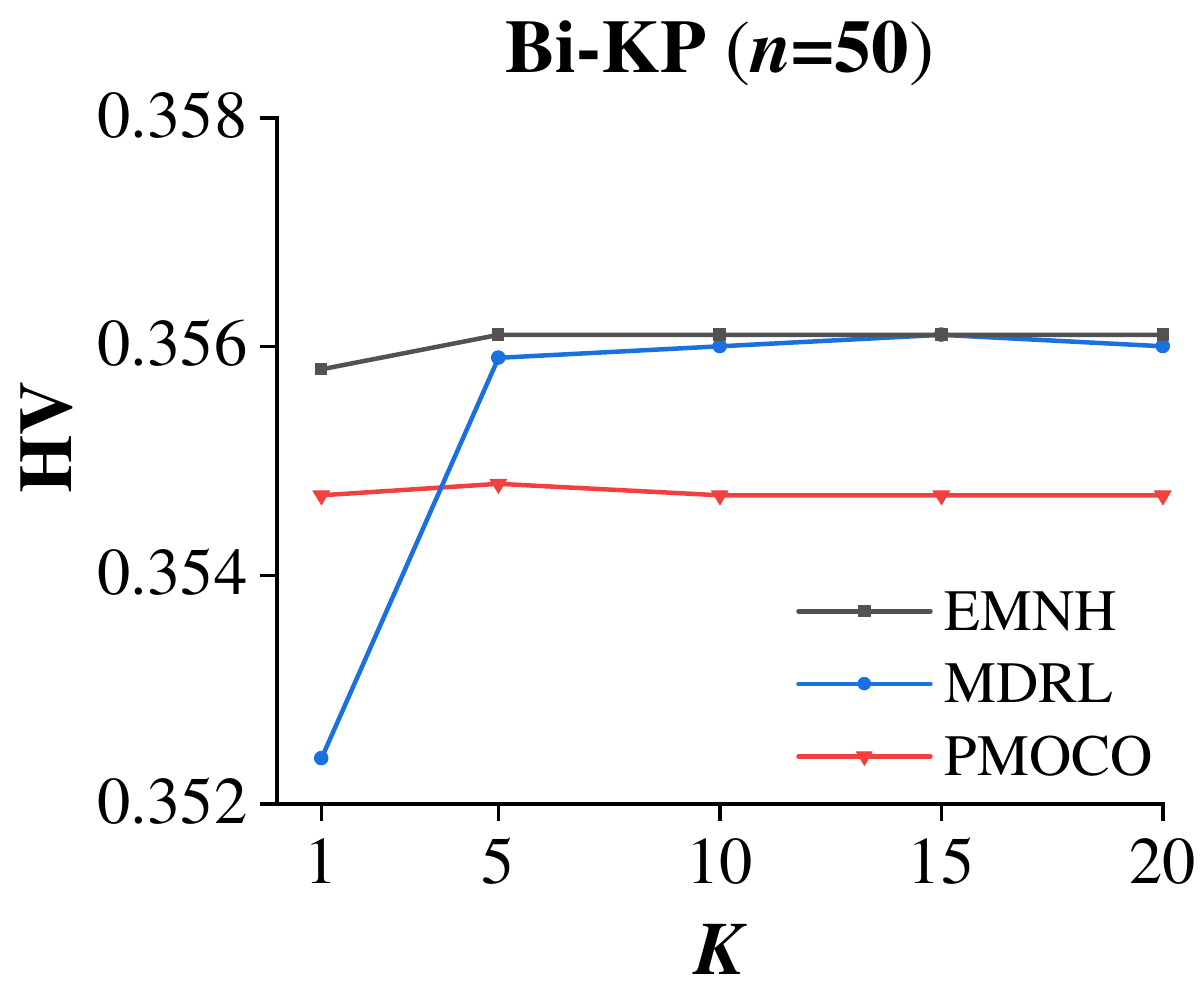}
			\label{figA5g}
		}
		\subfigure[]{
			\centering
			\includegraphics[width=0.315\textwidth]{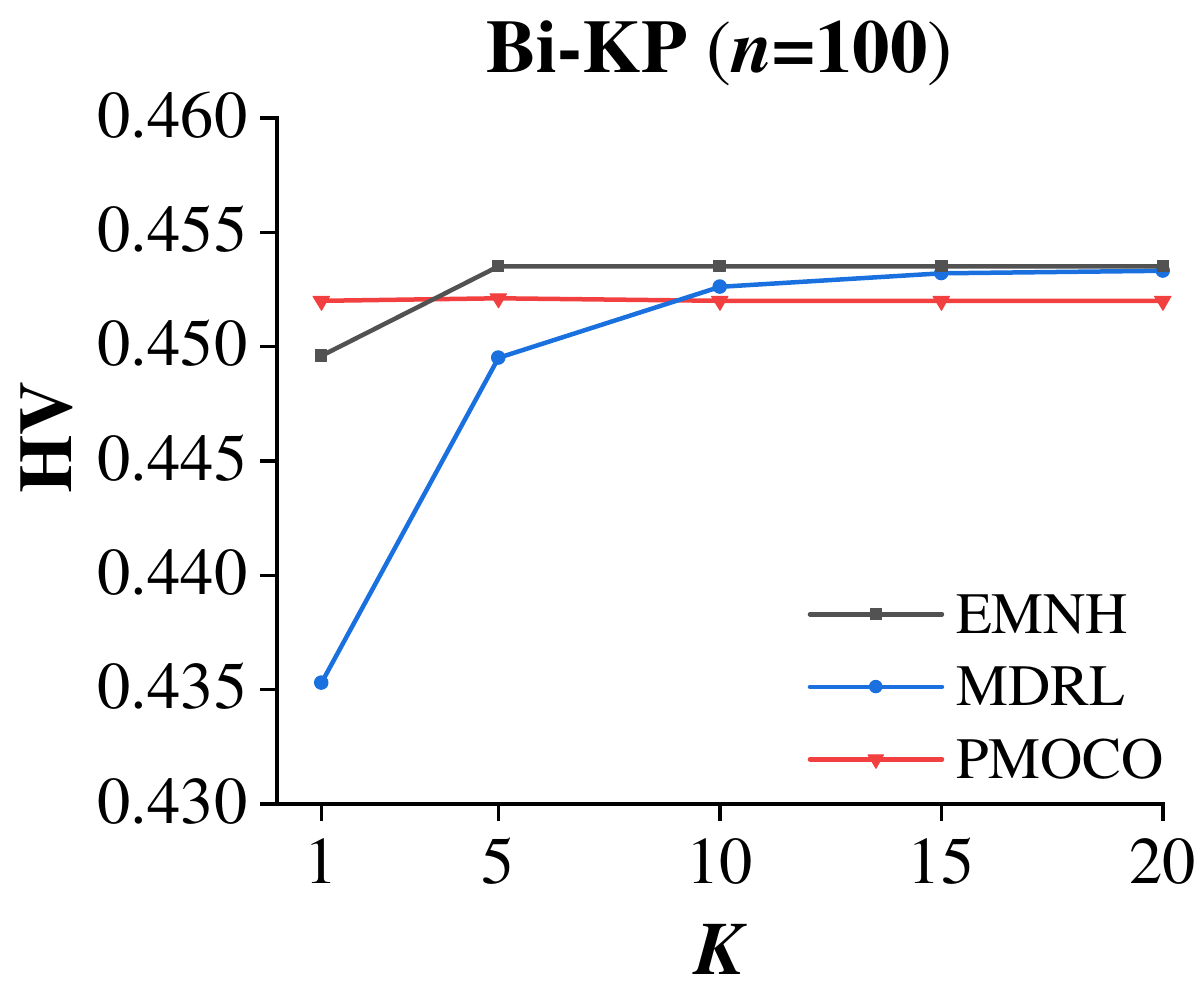}
			\label{figA5h}
		}
		\subfigure[]{
			\centering
			\includegraphics[width=0.315\textwidth]{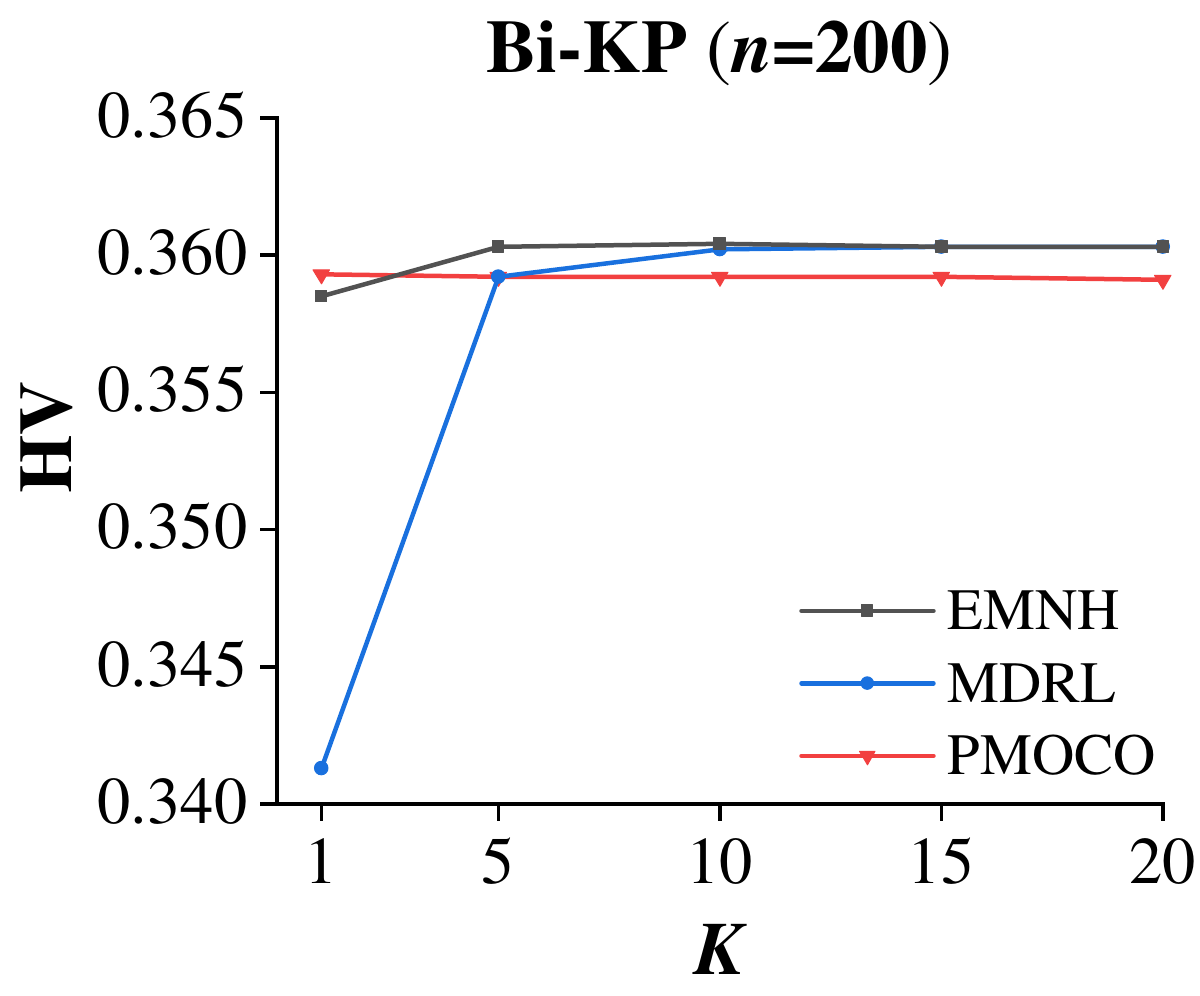}
			\label{figA5i}
		}
		\subfigure[]{
			\centering
			\includegraphics[width=0.315\textwidth]{fig/figA5j.pdf}
			\label{figA5j}
		}
		\subfigure[]{
			\centering
			\includegraphics[width=0.315\textwidth]{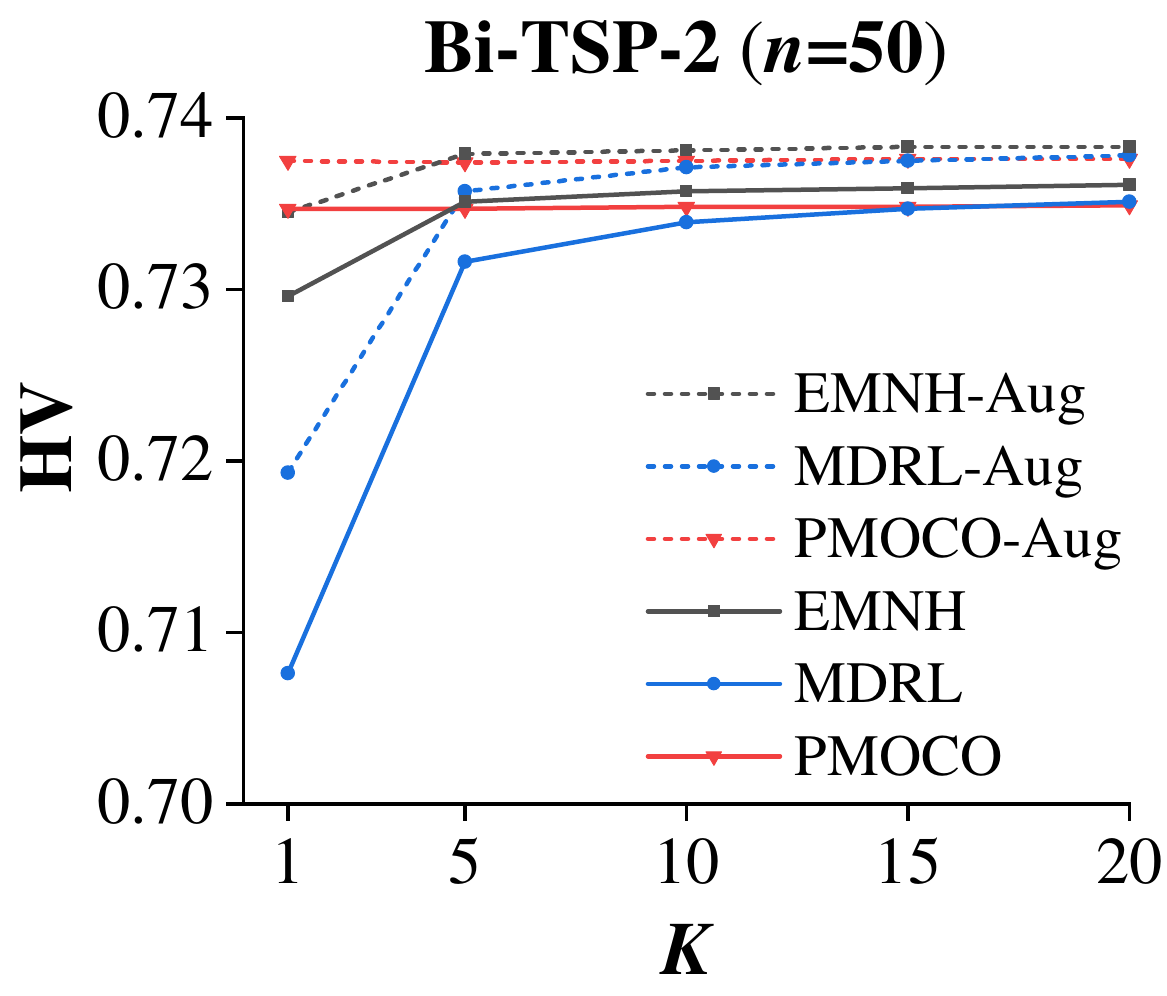}
			\label{figA5k}
		}
		\subfigure[]{
			\centering
			\includegraphics[width=0.315\textwidth]{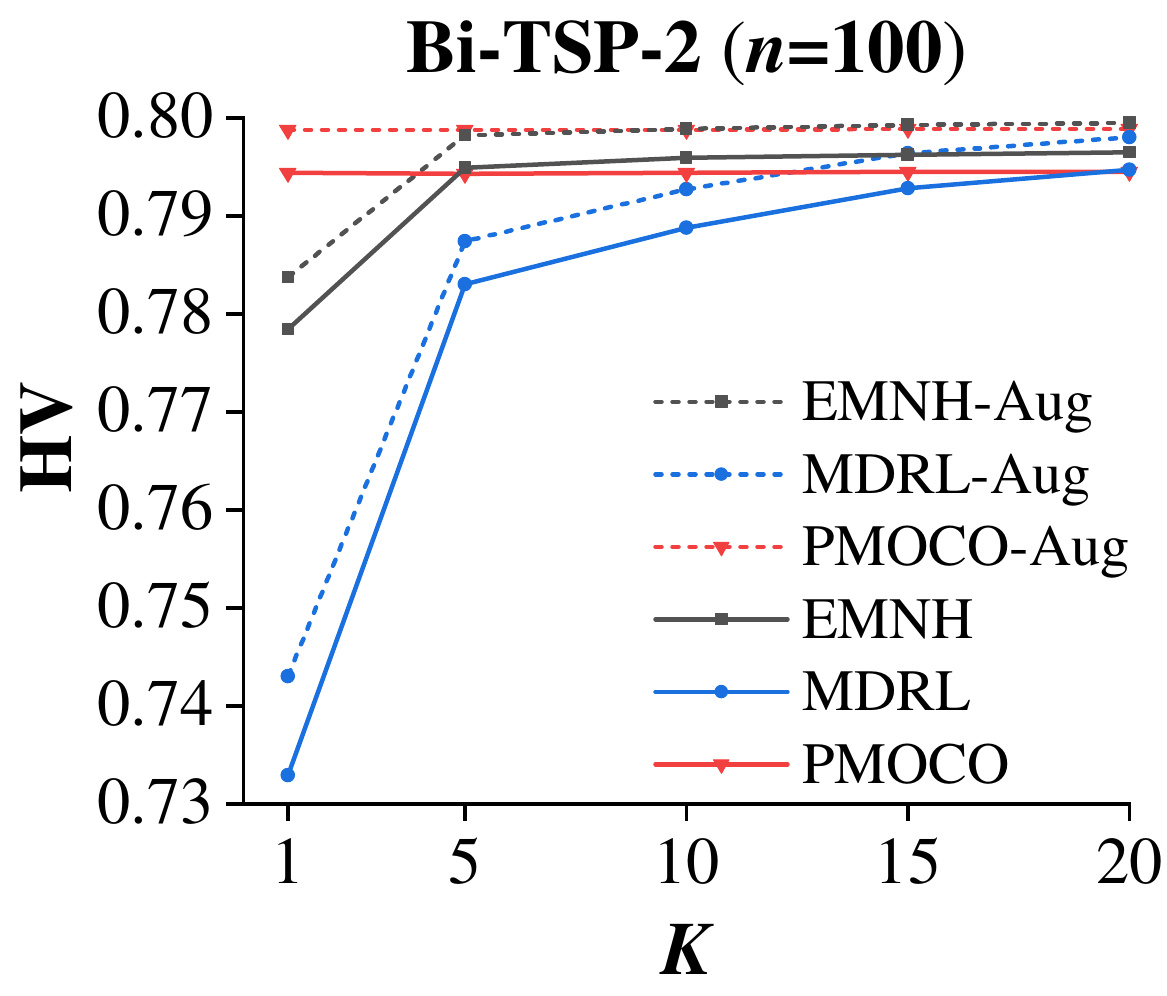}
			\label{figA5l}
		}
		\caption{Fine-tuning efficiency on various MOCOPs.}
		\label{figA5}
	\end{figure}

	Figure \ref{figA5} shows that, on various MOCOPs, EMNH attains better solution quality than MDRL with equal total fine-tuning steps, especially a larger gap with a few fine-tuning steps, which means that EMNH has higher fine-tuning efficiency. Furthermore, PMOCO is equipped with the hierarchical fine-tuning method for a fair comparison. However, it could hardly get improved, which indicates that EMNH has more potential to derive performant submodels to deal with specific tasks. Besides, in most cases, EMNH outperforms PMOCO only with a few fine-tuning steps ($K=5$).
	
	\section{Instance augmentation}
	
	Instance augmentation exploits multiple efficient transformations for the original instance that share the same optimal solution. Then, all transformed problems are solved and the best solution among them are finally selected. According to POMO \cite{kwo20}, a 2D coordinate $(x,y)$ has eight different transformations, \{$(x,y),(y,x),(x,1-y),(y,1-x),(1-x,y),(1-y,x),(1-x,1-y),(1-y,1-x)$\}.
	
	For the $M$-objective TSP type 1, each node has $M$ 2D coordinates, so it has $8^M$ different transformations. For the $M$-objective TSP type 2, each node has $(M-1)$ 2D coordinates and a 1D coordinate. The 1D coordinate $x$ has two different transformations, \{$x,1-x$\}. Thus, the $M$-objective TSP type 2 has $2 \times 8^{M-1}$ different transformations. For MOCVRP, each node has a 2D coordinate, so it has $8$ different transformations. MOKP has no transformation.
	
	\section{More results on MOCOPs with imbalanced objective domains}

    \subsection{Scaled weight sampling method for training}
	
	For the problems with imbalanced objective domains, including Bi-CVRP, Bi-TSP-2, and Tri-TSP-2, we further study PMOCO with a scaled sampling method in the training phase, denoted as PMOCO-S. Specifically, in each sampling for PMOCO, a sampled weight vector $\bm{\lambda}$ is sacled by $\bm{f}'$ as $\lambda^s_m=\lambda_m/f'_m$, where $f'_m$ is dynamically estimated by the model on a validation dataset associated with $\bm{\lambda}=(1/M,\dots,1/M)$ during training. Table \ref{tabA3} shows that PMOCO-S could certainly improve the performance of PMOCO for Bi-CVRP and Tri-TSP-2, but it is still inferior to EMNH.
	
	\begin{table}[!t]
		\caption{PMOCO with the scaled sampling method.}
		\centering
		\begin{tabular}{cc|cc|cc|cc}
			\toprule
			&       & \multicolumn{2}{c|}{$n$=20} & \multicolumn{2}{c|}{$n$=50} & \multicolumn{2}{c}{$n$=100} \\
			& Method & HV$\uparrow$ & Gap$\downarrow$   & HV$\uparrow$ & Gap$\downarrow$   & HV$\uparrow$ & Gap$\downarrow$ \\
			\midrule
			\multirow{6}*{\rotatebox{90}{Bi-CVRP}} & PMOCO & 0.4267  & 0.81\% & 0.4036  & 1.70\% & 0.3913  & 4.07\% \\
			& PMOCO-S & 0.4274  & 0.65\% & 0.4057  & 1.19\% & 0.4042  & 0.91\% \\
			& EMNH & \underline{0.4299}  & \underline{0.07\%} & \underline{0.4098}  & \underline{0.19\%} & \underline{0.4072}  & \underline{0.17\%} \\
			& PMOCO-Aug & 0.4294  & 0.19\% & 0.4080  & 0.63\% & 0.3969  & 2.70\% \\
			& PMOCO-S-Aug & 0.4295  & 0.16\% & 0.4090  & 0.39\% & 0.4063  & 0.39\% \\
			& EMNH-Aug & \textbf{0.4302} & \textbf{0.00\%} & \textbf{0.4106} & \textbf{0.00\%} & \textbf{0.4079} & \textbf{0.00\%} \\
			\midrule
			\multirow{6}*{\rotatebox{90}{Bi-TSP-2}} & PMOCO & 0.6590  & 1.18\% & 0.7347  & 0.51\% & 0.7944  & 0.65\% \\
			& PMOCO-S & 0.6520  & 2.23\% & 0.7333  & 0.70\% & 0.7927  & 0.86\% \\
			& EMNH & \textbf{0.6669} & \textbf{0.00\%} & 0.7361  & 0.32\% & 0.7965  & 0.39\% \\
			& PMOCO-Aug & \underline{0.6653}  & \underline{0.24\%} & \underline{0.7375}  & \underline{0.14\%} & \underline{0.7988}  & \underline{0.10\%} \\
			& PMOCO-S-Aug & 0.6624  & 0.67\% & 0.7366  & 0.26\% & 0.7974  & 0.28\% \\
			& EMNH-Aug & \textbf{0.6669} & \textbf{0.00\%} & \textbf{0.7385} & \textbf{0.00\%} & \textbf{0.7996} & \textbf{0.00\%} \\
			\midrule
			\multirow{6}*{\rotatebox{90}{Tri-TSP-2}} & PMOCO & 0.5020  & 0.30\% & 0.5176  & 1.95\% & 0.5777  & 2.02\% \\
			& PMOCO-S & 0.4917  & 2.34\% & 0.5175  & 1.97\% & 0.5778  & 2.00\% \\
			& EMNH & 0.5022  & 0.26\% & 0.5205  & 1.40\% & 0.5813  & 1.41\% \\
			& PMOCO-Aug & \textbf{0.5035} & \textbf{0.00\%} & 0.5258  & 0.40\% & 0.5862  & 0.58\% \\
			& PMOCO-S-Aug & \underline{0.5029}  & \underline{0.12\%} & \underline{0.5259}  & \underline{0.38\%} & \underline{0.5863}  & \underline{0.56\%} \\
			& EMNH-Aug & \textbf{0.5035} & \textbf{0.00\%} & \textbf{0.5279} & \textbf{0.00\%} & \textbf{0.5896} & \textbf{0.00\%} \\
			\bottomrule
		\end{tabular}%
		\label{tabA3}%
	\end{table}%

    \subsection{Scaled weight assignment method for inference}

    Similar to other decomposition-based methods like PMOCO and MDRL, EMNH also faces the challenge of non-uniformly solution distributions. However, this issue can be mitigated by employing appropriate weight assignment methods during inference, as EMNH offers the flexibility to handle arbitrary weight vectors. Specifically, when the approximate scales of different objectives are known, we can normalize them to [0,1] to achieve a more uniform Pareto front. Alternatively, we can adjust the weight assignment to generate a more uniform Pareto front.

    A scaled weight assignment (SWA) method can be directly applied to alleviate this issue. Specifically, each uniform weight vector $\bm{\lambda}$ is scaled by $\bm{f}'$ as $\lambda^s_m=\lambda_m/f'_m$ and normalized to $[0,1]^M$. Here, $\bm{f}'$ is estimated using a validation dataset associated with $\bm{\lambda}=(1/M,\dots,1/M)$. The advantage of this SWA method is that it does not require prior problem information.

\begin{figure}[!t]
	\centering
	\subfigure[]{
		\centering
		\includegraphics[width=0.31\textwidth]{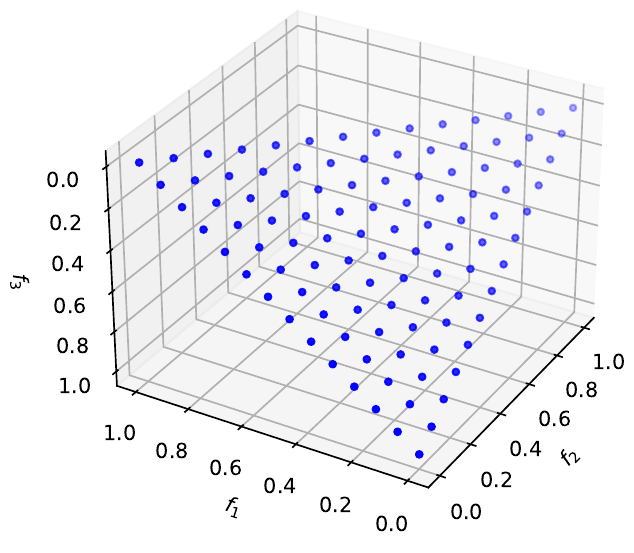}
	}
	\subfigure[]{
		\centering
		\includegraphics[width=0.31\textwidth]{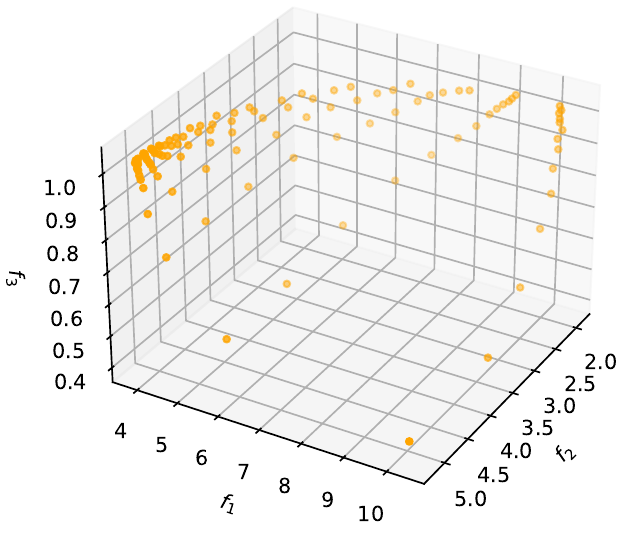}
	}
	\subfigure[]{
		\centering
		\includegraphics[width=0.31\textwidth]{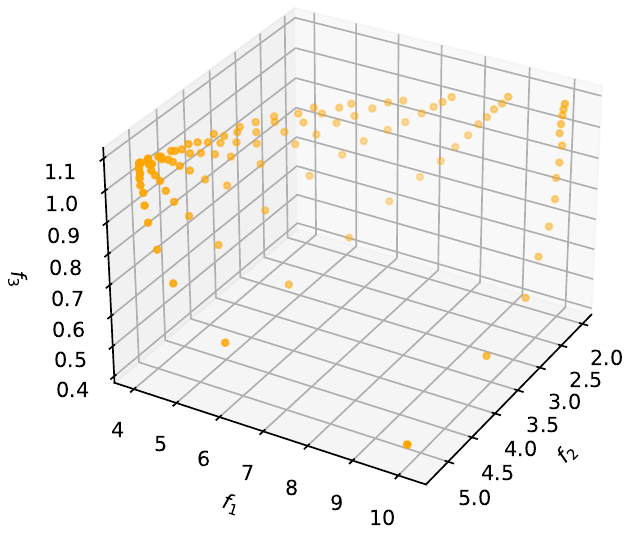}
	}
	\subfigure[]{
		\centering
		\includegraphics[width=0.31\textwidth]{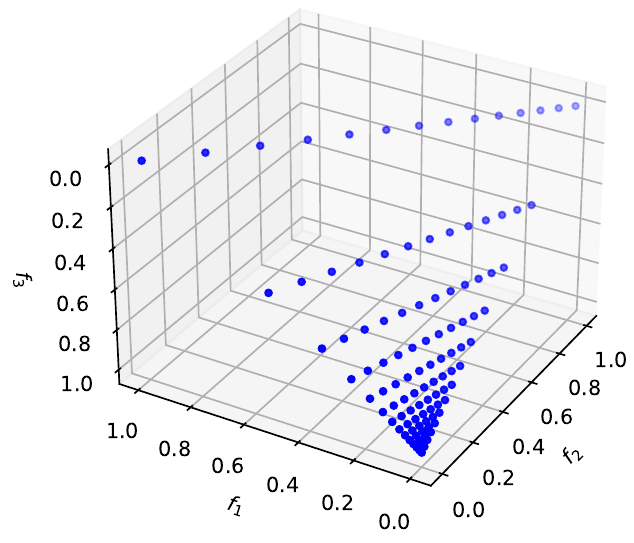}
	}
	\subfigure[]{
		\centering
		\includegraphics[width=0.31\textwidth]{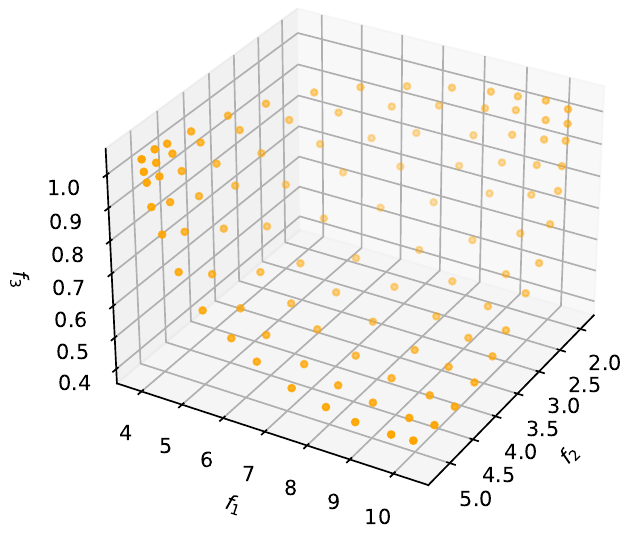}
	}
	\subfigure[]{
		\centering
		\includegraphics[width=0.31\textwidth]{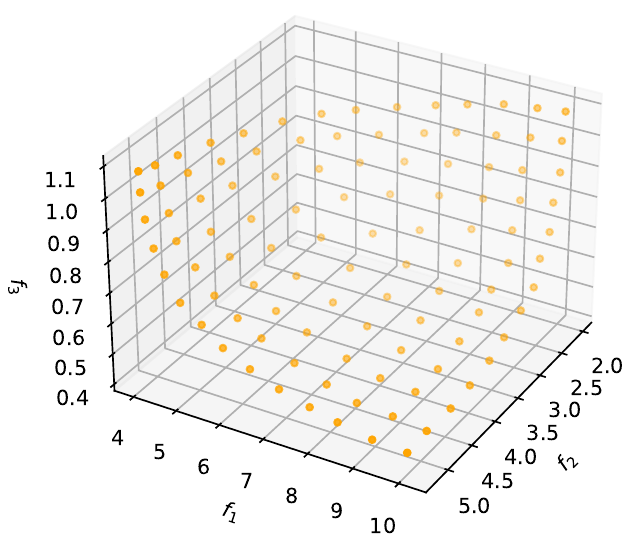}
	}
	\subfigure[]{
		\centering
		\includegraphics[width=0.31\textwidth]{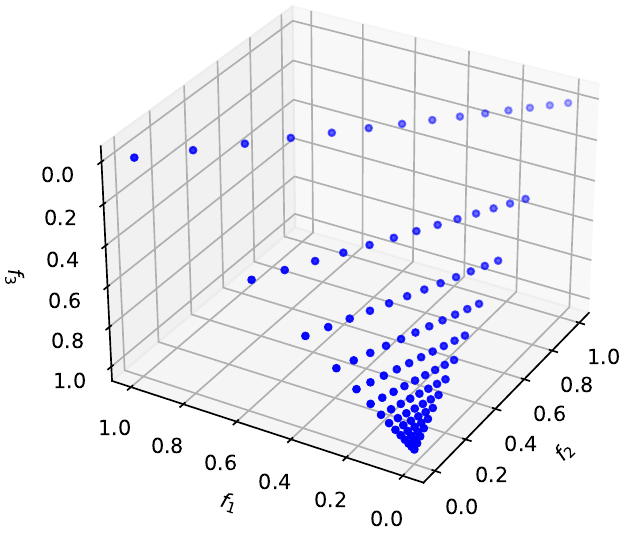}
	}
	\subfigure[]{
		\centering
		\includegraphics[width=0.31\textwidth]{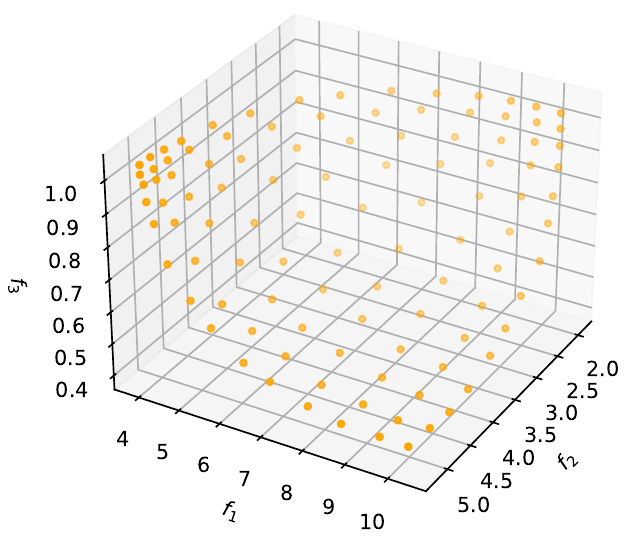}
	}
	\caption{Solutions generated by using 105 uniform/non-uniform distributed weights on Tri-TSP-1 ($n$=20) with asymmetric Pareto front. (a) Uniform weights. (b) EMNH with uniform weights. (c) PMOCO with uniform weights. (d) A priori non-uniform weights. (e) EMNH with a priori non-uniform weights. (f) PMOCO with a priori non-uniform weights. (g) Weights generated by the scaled weight assignment (SWA) method. (h) EMNH with SWA.}
	\label{fig:swa}
    \end{figure}

    The results on Tri-TSP instances with asymmetric Pareto fronts are presented in Figure \ref{fig:swa}. For these instances, the coordinates for the three objectives are randomly sampled from $[0,1]^2$, $[0,0.5]^2$, $[0,0.1]^2$, respectively. The results demonstrate that EMNH-SWA effectively produces a more uniform Pareto front. Compared to a scaling weight method with prior knowledge used in PMOCO \cite{lin22}, where uniform weight vectors are element-wise multiplied by (1,2,10) and then normalized back to $[0,1]^3$, EMNH-SWA achieves desirable performance.

	\section{Detailed results of generalization capability}
	
	We test the generalization capability of our EMNH on larger-scale random instances ($n$=150/200) and three commonly used MOTSP benchmark instances (KroAB100/150/200) in TSPLIB \cite{rei91}, as shown in Tables \ref{tabAA1} and \ref{tabA1}, respectively. EMNH, which is trained and fine-tuned both on the instances with $n$=100, has a superior generalization capability compared with the state-of-the-art MOEA and other neural heuristics for larger problem sizes.
	
	\begin{table}[!t]
		\caption{Results of generalization capability on 200 random instances .}
		\centering
		\begin{tabular}{c|ccc|ccc}
			\toprule
			& \multicolumn{3}{c|}{Bi-TSP-1 ($n$=150)} & \multicolumn{3}{c}{Bi-TSP-1 ($n$=200)} \\
			Method & HV$\uparrow$ & Gap$\downarrow$   & Time  & HV$\uparrow$ & Gap$\downarrow$   & Time \\
			\midrule
			WS-LKH & \textbf{0.7149} & \textbf{-2.38\%} & 13h   & \textbf{0.7490} & \textbf{-2.50\%} & 22h \\
			PPLS/D-C & 0.6784  & 3.25\% & 21h   & 0.7106  & 3.11\% & 32h \\
			\midrule
			DRL-MOA & 0.6901  & 1.17\% & 45s   & 0.7219  & 1.20\% & 87s \\
			PMOCO & 0.6910  & 1.05\% & 45s   & 0.7231  & 1.04\% & 87s \\
			MDRL  & 0.6922  & 0.87\% & 40s   & 0.7251  & 0.77\% & 84s \\
			EMNH & 0.6930  & 0.76\% & 40s   & 0.7260  & 0.64\% & 83s \\
			PMOCO-Aug & 0.6967  & 0.23\% & 47m   & 0.7283  & 0.33\% & 1.6h \\
			MDRL-Aug & 0.6976  & 0.10\% & 47m   & 0.7299  & 0.11\% & 1.6h \\
			EMNH-Aug & \underline{0.6983}  & \underline{0.00\%} & 47m   & \underline{0.7307}  & \underline{0.00\%} & 1.6h \\
			\bottomrule
		\end{tabular}%
		\label{tabAA1}%
	\end{table}%

	\begin{table}[!t]
		\caption{Results of generalization capability on benchmark instances.}
		\centering
            \small
		\begin{tabular}{c|ccc|ccc|ccc}
			\toprule
			& \multicolumn{3}{c|}{KroAB100} & \multicolumn{3}{c|}{KroAB150} & \multicolumn{3}{c}{KroAB200} \\
			Method & HV$\uparrow$ & Gap$\downarrow$   & Time  & HV$\uparrow$ & Gap$\downarrow$   & Time  & HV$\uparrow$ & Gap$\downarrow$   & Time \\
			\midrule
			WS-LKH & \textbf{0.7022} & \textbf{-0.92\%} & 2.3m  & \textbf{0.7017} & \textbf{-1.81\%} & 4.0m  & \textbf{0.7430} & \textbf{-2.20\%} & 5.6m \\
			PPLS/D-C & 0.6785  & 2.77\% & 31m   & 0.6659  & 3.84\% & 1.1h  & 0.7100  & 2.69\% & 3.1h \\
			\midrule
			DRL-MOA & 0.6903  & 0.79\% & 10s   & 0.6794  & 1.42\% & 18s   & 0.7185  & 1.17\% & 23s \\
			PMOCO & 0.6878  & 1.15\% & 9s    & 0.6819  & 1.06\% & 17s   & 0.7193  & 1.06\% & 23s \\
			MDRL  & 0.6881  & 1.11\% & 10s    & 0.6831  & 0.89\% & 17s   & 0.7209  & 0.84\% & 23s \\
			EMNH & 0.6900  & 0.83\% & 9s    & 0.6832  & 0.87\% & 16s   & 0.7217  & 0.73\% & 23s \\
			PMOCO-Aug & 0.6937  & 0.30\% & 12s    & 0.6886  & 0.09\% & 19s   & 0.7251  & 0.26\% & 27s \\
			MDRL-Aug & 0.6950  & 0.11\% & 13s    & 0.6890  & 0.03\% & 19s   & 0.7261  & 0.12\% & 28s \\
			EMNH-Aug & \underline{0.6958} & \underline{0.00\%} & 12s    & \underline{0.6892}  & \underline{0.00\%} & 18s   & \underline{0.7270}  & \underline{0.00\%} & 27s \\
			\bottomrule
		\end{tabular}%
		\label{tabA1}%
	\end{table}%

	\section{Hyper-parameter study}
	
	\subsection{Number of sampled weight vectors}
	
\begin{figure}[!t]
		\centering
		\subfigure[]{
			\centering
			\includegraphics[width=0.48\textwidth]{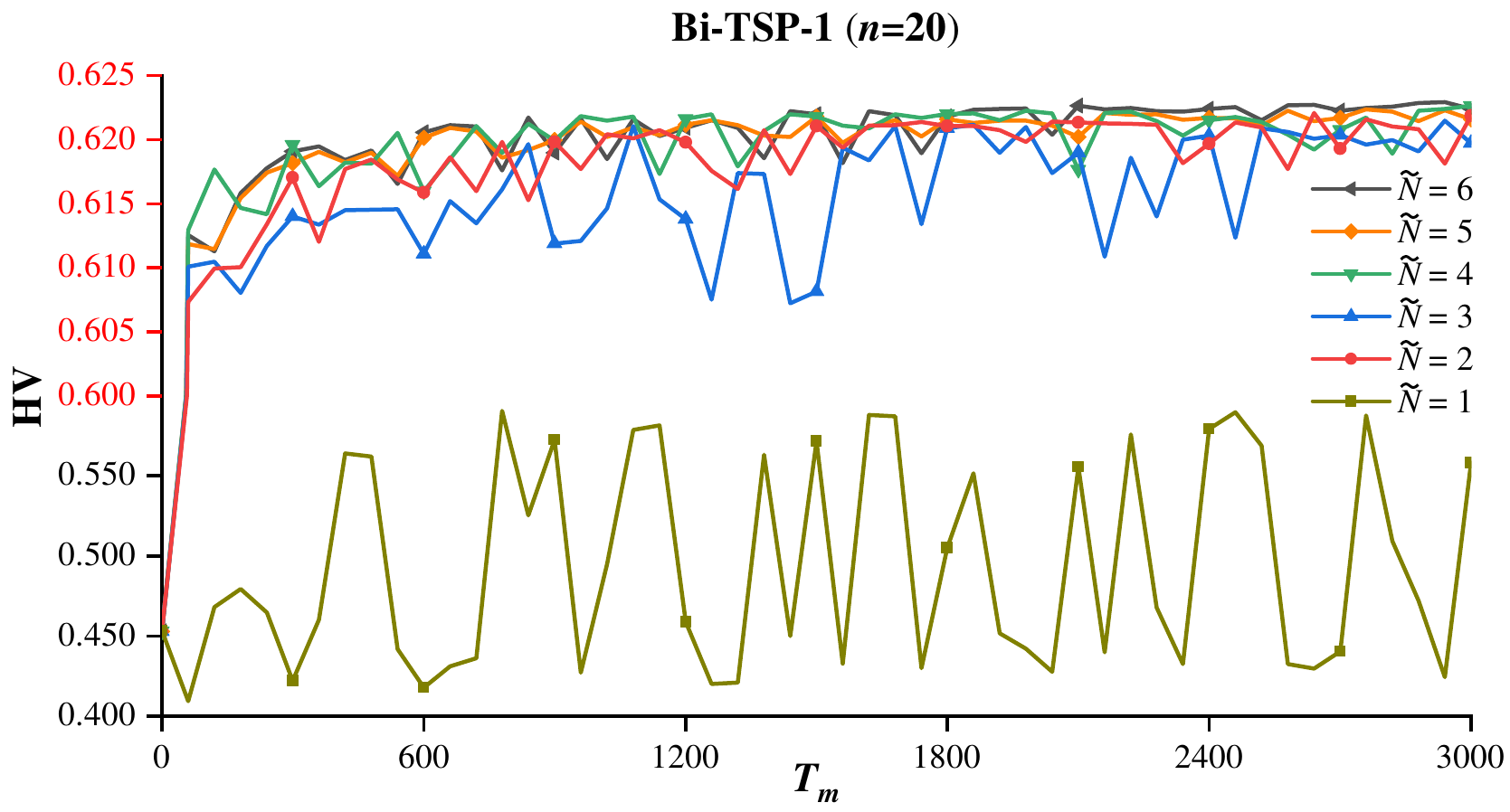}
			\label{figA6a}
		}
		\subfigure[]{
			\centering
			\includegraphics[width=0.48\textwidth]{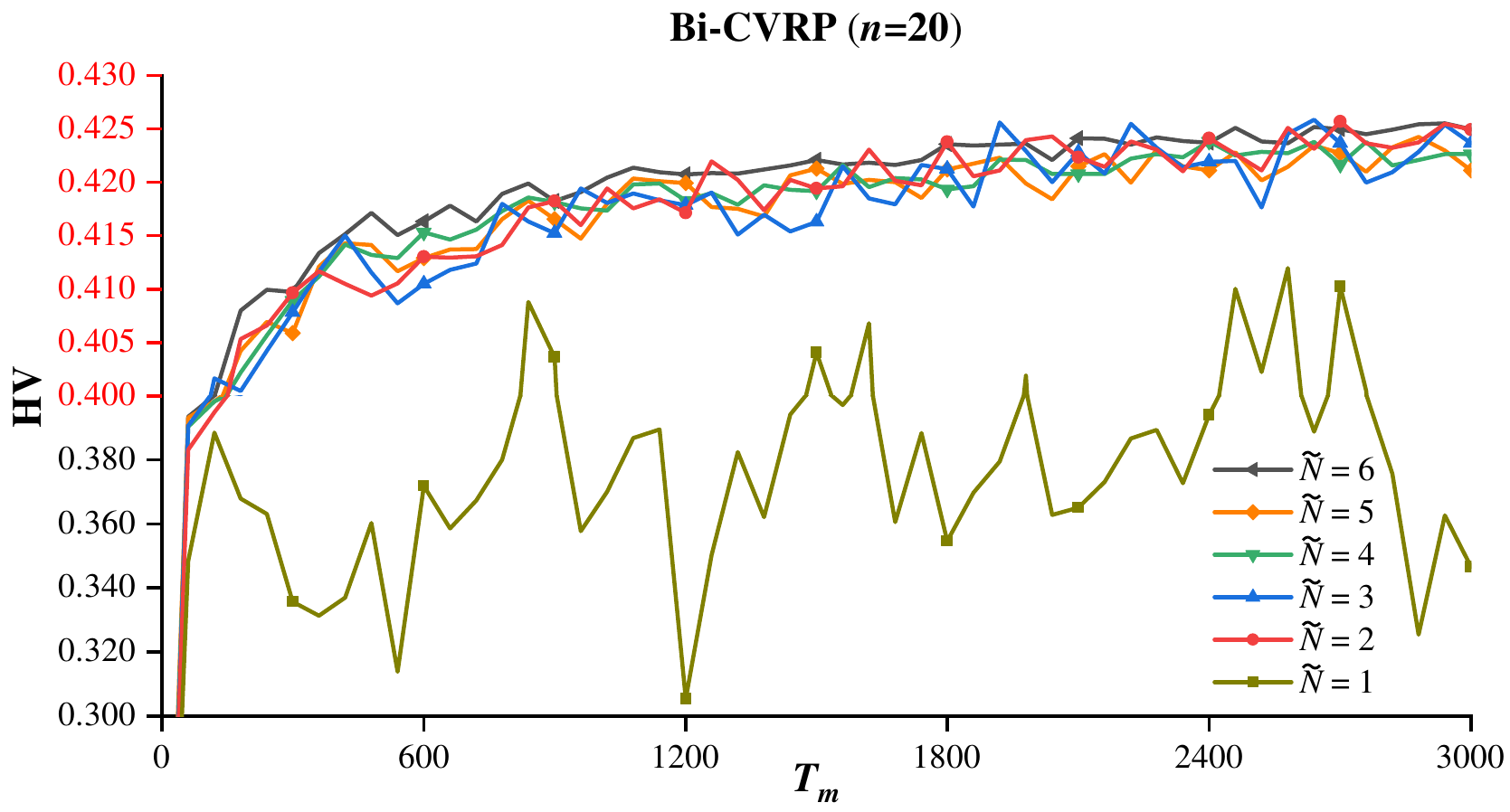}
			\label{figA6b}
		}
		\caption{Effects of the number of sampled weight vectors.}
		\label{figA6}
	\end{figure}

	Figure \ref{figA6} shows the results for various $\tilde{N}$. For bi-objective problems, $\tilde{N}=1$ leads to a significantly unstable training process, since the sample number is quite small and it has no symmetric sample. $\tilde{N}=3$ also causes a slightly unstable training process. $\tilde{N}=kM$ with $k \in \{1,2,\dots\}$ can effectively stabilize the training process. In summary, $\tilde{N}=M$ is a more favorable setting.
	
	\subsection{Scalarization method}
	
	EMNH is generic for solving MOCOPs based on decomposition \cite{zha07}, which can employ various scalarization methods, including weighted sum (WS) and Tchebycheff (TCH).
	
	WS is the simplest method and is effective for the convex $\mathcal{PF}$. It considers the linear combination of $M$ objectives, as follows,
	\begin{equation}
		\underset{x \in \mathcal{X}}{\min}\; g_{\rm{ws}}(x|\bm{\lambda}) = \sum_{m=1}^{M}\lambda_m f_m(x).
	\end{equation}
	 In theory, TCH can tackle the concave $\mathcal{PF}$, but it would lead to a more complex objective function. It is defined as follows,
	 \begin{equation}
	 	\underset{x \in \mathcal{X}}{\min}\; g_{\rm{tch}}(x|\bm{\lambda}) = \underset{1\leq m\leq M}{\max}\; \{\lambda_m|f_m(x)-z^*_m|\},
	 \end{equation}
	 where $z^*_m<\min_{x \in \mathcal{X}}f_m(x)$ is an ideal value of $f_m(x)$.
	 
	As shown in Table \ref{tabA4}, the results show that WS is a simple yet effective scalarization method for the studied problems. In principle, EMNH can freely use any existing scalarizing function. Different scalarization methods have their own strengths and drawbacks, but the study with respect to scalarization methods is beyond the scope of this paper, which we will investigate in the future.

	\begin{table}[!t]
		\caption{Effects of the scalarization method.}
		\centering
		\begin{tabular}{cc|cc|cc|cc}
			\toprule
			&       & \multicolumn{2}{c|}{$n$=20} & \multicolumn{2}{c|}{$n$=50} & \multicolumn{2}{c}{$n$=100} \\
			& Method & HV$\uparrow$ & Gap$\downarrow$   & HV$\uparrow$ & Gap$\downarrow$   & HV$\uparrow$ & Gap$\downarrow$ \\
			\midrule
			\multirow{4}[2]{*}{\rotatebox{90}{Bi-TSP-1}} & EMNH-TCH & \underline{0.6271}  & \underline{0.00\%} & 0.6331  & 1.20\% & 0.6927  & 1.37\% \\
			& EMNH-WS & \underline{0.6271}  & \underline{0.00\%} & 0.6364  & 0.69\% & 0.6969  & 0.77\% \\
			& EMNH-TCH-Aug & \textbf{0.6294} & \textbf{-0.37\%} & \underline{0.6401}  & \underline{0.11\%} & \underline{0.6995}  & \underline{0.40\%} \\
			& EMNH-WS-Aug & \underline{0.6271}  & \underline{0.00\%} & \textbf{0.6408} & \textbf{0.00\%} & \textbf{0.7023} & \textbf{0.00\%} \\
			\midrule
			\multirow{4}[2]{*}{\rotatebox{90}{Tri-TSP-1}} & EMNH-TCH & 0.4665  & 1.00\% & 0.4175  & 5.50\% & 0.4681  & 5.87\% \\
			& EMNH-WS & 0.4698  & 0.30\% & \underline{0.4324}  & \underline{2.13\%} & \underline{0.4866}  & \underline{2.15\%} \\
			& EMNH-TCH-Aug & \underline{0.4710}  & \underline{0.04\%} & 0.4295  & 2.78\% & 0.4814  & 3.20\% \\
			& EMNH-WS-Aug & \textbf{0.4712} & \textbf{0.00\%} & \textbf{0.4418} & \textbf{0.00\%} & \textbf{0.4973} & \textbf{0.00\%} \\
			\bottomrule
		\end{tabular}%
		\label{tabA4}%
	\end{table}%

    \section{Trade-off between lightweight fine-tuning and performance}

    For a given weight vector, EMNH fine-tunes the meta-model to derive a submodel to solve the corresponding subproblem. We study another two (relatively) lightweight fine-tuning methods, including only updating the head parameter (denoted as EMNH-FH) according to feature reuse \cite{rag20} and only updating the decoder parameter (denoted as EMNH-FD) like PMOCO \cite{lin22}. These two methods even allow us to only fine-tune and store parts of the original submodels, i.e., $N$ heads or $N$ decoders, thereby being more computationally efficient. Meanwhile, such benefit may bring about performance sacrifices in some cases. We report the results in Table \ref{tab:light} and the parameter numbers of various parts of the model in Table \ref{tab:param}. The lightweight fine-tuning has slightly inferior performance compared with the original EMNH in most cases except on Bi-CVRP ($n$=100). Generally, the more lightweight of the fine-tuning, the more performance deterioration (i.e.,EMNH-FH v.s. EMNH-FD as displayed in the table below, where FH is more light than FD). However, these lightweight fine-tuning methods can be used as alternatives when the computational and memory resources are limited.

    \begin{table}[!t]
      \centering
      \caption{Results of lightweight fine-tuning methods.}
        \begin{tabular}{cc|cc|cc}
        \toprule
              &       & \multicolumn{2}{c|}{$n$=20} & \multicolumn{2}{c}{$n$=100} \\
              & Method & HV    & Gap   & HV    & Gap \\
        \midrule
        \multirow{5}[2]{*}{\rotatebox{90}{Bi-CVRP}} & PMOCO-Aug & 0.4294  & 0.19\% & 0.3966  & 2.77\% \\
              & MDRL-Aug & 0.4292  & 0.23\% & 0.4072  & 0.17\% \\
              & EMDRL-Aug & 0.4302  & 0.00\% & 0.4079  & 0.00\% \\
              & EMDRL-FD-Aug & 0.4299  & 0.07\% & 0.4082  & -0.07\% \\
              & EMDRL-FH-Aug & 0.4298  & 0.09\% & 0.4082  & -0.07\% \\
        \midrule
        \multirow{5}[2]{*}{\rotatebox{90}{Tri-TSP-1}} & PMOCO-Aug & 0.4712  & 0.00\% & 0.4956  & 0.34\% \\
              & MDRL-Aug & 0.4712  & 0.00\% & 0.4958  & 0.30\% \\
              & EMDRL-Aug & 0.4712  & 0.00\% & 0.4973  & 0.00\% \\
              & EMDRL-FD-Aug & 0.4710  & 0.04\% & 0.4925  & 0.97\% \\
              & EMDRL-FH-Aug & 0.4707  & 0.11\% & 0.4906  & 1.35\% \\
        \bottomrule
        \end{tabular}%
      \label{tab:light}%
    \end{table}%

    \begin{table}[!t]
      \centering
      \caption{Parameter numbers of various parts of models.}
        \begin{tabular}{l|ccc|ccc}
        \toprule
              & \multicolumn{3}{c|}{Bi-CVRP Model	} & \multicolumn{3}{c}{Tri-TSP-1 Model	} \\
              & Whole Model & Decoder & Head  & Whole Model & Decoder & Head \\
        \midrule
        \#(Parameters) & 1287K & 98K   & 16K   & 1303K & 115K  & 16K \\
        \bottomrule
        \end{tabular}%
      \label{tab:param}%
    \end{table}%

\begin{figure}[!t]
	\centering
	\subfigure[]{
		\centering
		\includegraphics[width=0.31\textwidth]{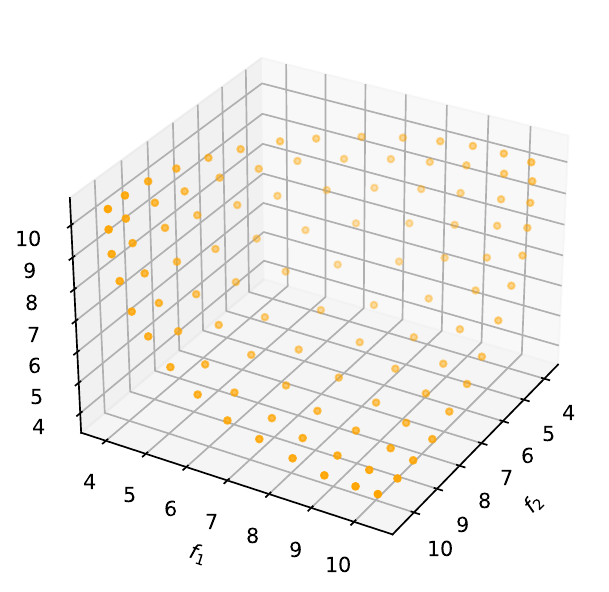}
	}
	\subfigure[]{
		\centering
		\includegraphics[width=0.31\textwidth]{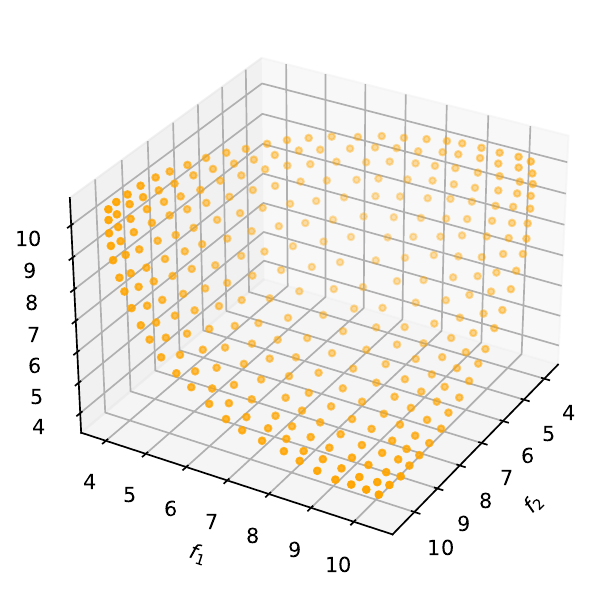}
	}
	\subfigure[]{
		\centering
		\includegraphics[width=0.31\textwidth]{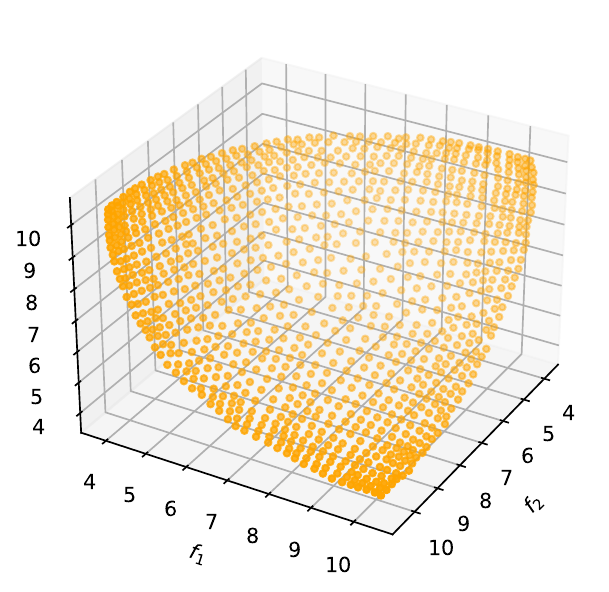}
	}
	\caption{Solutions of Tri-TSP-1 ($n$=20) generated with various numbers of weight vectors. (a) 105 weights. (b) 300 weights. (c) 1035 weights.}
	\label{fig:fh}
    \end{figure}

    Moreover, same as EMNH, EMNH-FH can also generate much more dense Pareto solutions to improve the performance via increasing weight vectors and corresponding fine-tuned heads. We have plotted the generated Pareto fronts with 105, 300 and 1035 weight vectors on Tri-TSP-1 which verified the above point, as shown in Figure \ref{fig:fh}.

\end{document}